%% file: arxiv.tex
\documentclass[sigconf]{acmart}
\usepackage{makecell}
\usepackage{xcolor}
\usepackage{multirow}
\usepackage{enumitem}
\usepackage[most]{tcolorbox}
\usepackage{tikz}

\AtBeginDocument{%
  }

\setcopyright{acmlicensed}
\copyrightyear{2018}
\acmYear{2018}
\acmDOI{XXXXXXX.XXXXXXX}
\acmConference[Conference acronym 'XX]{Make sure to enter the correct
  conference title from your rights confirmation email}{June 03--05,
  2018}{Woodstock, NY}
\acmISBN{978-1-4503-XXXX-X/2018/06}

\input{macros}

\begin{document}

\title{RouteHijack: Routing-Aware Attack on Mixture-of-Experts LLMs}

\author{Zhiyuan Xu}
\email{zhiyuan.xu@bristol.ac.uk}
\affiliation{%
  \institution{University of Bristol}
  \city{Bristol}
  \country{United Kingdom}
}

\author{Joseph Gardiner}
\email{joe.gardiner@bristol.ac.uk}
\affiliation{%
  \institution{University of Bristol}
  \city{Bristol}
  \country{United Kingdom}
}

\author{Sana Belguith}
\email{sana.belguith@bristol.ac.uk}
\affiliation{%
  \institution{University of Bristol}
  \city{Bristol}
  \country{United Kingdom}
}

\author{Lichao Wu}
\email{lichao.wu@bristol.ac.uk}
\affiliation{%
  \institution{University of Bristol}
  \city{Bristol}
  \country{United Kingdom}
}

\begin{abstract}
Safety alignment is critical for the responsible deployment of large language models (LLMs). As Mixture-of-Experts (MoE) architectures are increasingly adopted to scale model capacity, understanding their safety robustness becomes essential. Existing adversarial attacks, however, have notable limitations. Prompt-based jailbreaks rely on heuristic search and transfer poorly, model intervention methods require privileged access to internal representations, and optimization-based input attacks remain output-centric and are 
fundamentally limited to MoE models due to the non-differentiable routing mechanism.

In this paper, we present \ourname, a routing-aware jailbreak for MoE LLMs. Our key insight is that safety behavior is concentrated in a small subset of experts that are preferentially activated during refusal, creating an opportunity to steer model behavior by influencing routing decisions through input optimization. Building on this observation, \ourname first performs response-driven expert localization to identify safety-critical and harmful experts by contrasting activations under safe refusals and harmful completions. It then constructs adversarial suffixes with a routing-aware objective that suppresses safety experts, promotes harmful experts, and prevents early-stage refusal during generation. At inference time, the optimized suffix is appended to a malicious prompt, requiring only input access. Across seven state-of-the-art MoE LLMs, \ourname achieves a 69.3\% average attack success rate (ASR) and 89.1\% peak ASR, outperforming prior optimization-based attack by $3.2\times$ while incurring only a 1.3\% average utility drop across five NLU benchmarks. The optimized suffix also transfers zero-shot across five sibling MoE variants, raising average ASR from 27.7\% to 61.2\% across reasoning, human-preference, code, and general-knowledge settings, and further generalizes to three MoE-based VLMs, increasing average ASR from 2.47\% to 38.7\%. These findings expose a fundamental vulnerability in sparse expert architectures and highlight the need for defenses beyond output-level alignment.
\end{abstract}



\received{20 February 2007}
\received[revised]{12 March 2009}
\received[accepted]{5 June 2009}

\maketitle

\section{Introduction}
Large language models (LLMs) are increasingly deployed in a wide range of real-world applications, including content generation, medical assistance, and decision support systems~\cite{thirunavukarasu2023large,kaddour2023challenges}. Early advances were driven by dense Transformer-based models~\cite{vaswani2017attention}, where all parameters are activated for every token~\cite{kaplan2020scaling,chen2022towards}. Recent work has introduced Mixture-of-Experts (MoE) architectures, which enable efficient scaling by activating only a small subset of parameters per token~\cite{tian2025greaterleveragescalinglaws}. This sparsity enables substantially larger model capacity without proportional computational cost and has been widely adopted in state-of-the-art systems~\cite{openai_gpt54_2026,dai2024deepseekmoeultimateexpertspecialization, liu2024deepseek,qwen_moe_2024,qwen3.5_2026}. 

Alongside these architectural advances and capability improvement, ensuring the safety of LLMs has become a critical concern. Modern models are typically aligned using techniques such as supervised fine-tuning~\cite{qi2024safety} and reinforcement learning from human feedback~\cite{ouyang2022training}, which aim to suppress harmful or policy-violating outputs. However, safety-aligned LLMs remain vulnerable to adversarial attacks. 
For instance, input-level attacks, such as jailbreaks, rely on carefully-crafted prompts that exploit weaknesses in instruction following~\cite{wei2023jailbroken,niu2024jailbreaking,shen2024anything}. While popular and effective, these methods depend on heuristic design and trial-and-error search, making them brittle and difficult to transfer across models. In contrast, model-level interventions, such as model editing and parameter-level interventions, directly modify internal weights or activations to bypass safety mechanisms, but require access to model parameters or activations, which is unrealistic in most deployment scenarios~\cite{lai2025safexanalyzingvulnerabilitiesmoebased, wang2025badmoebackdooringmixtureofexpertsllms,wu2025gatebreaker,lintelo2026large, jiang2026sparse}.

\noindent\textbf{Optimization-based Attacks and Their Blind Spots.}
To bridge the gap between input-level and model-level attacks, recent works have proposed optimization-based jailbreak methods such as Greedy Coordinate Gradient (GCG)~\cite{zou2023universal,liu2023autodan,liao2024amplegcg}. GCG iteratively refines an adversarial suffix to increase the likelihood of a target response prefix (e.g., ``Sure, here is...''). By incorporating gradient information, these methods partially leverage model internals while remaining input-only at deployment time.
However, optimization-based attacks remain inherently output-centric and fail to account for the internal structure of modern architectures, particularly in modern Mixture-of-Experts (MoE) models. First, optimization-based methods are developed for dense LLMs and do not account for sparse routing. In MoE architectures, the Top-K routing operation introduces non-differentiability, disrupting gradient propagation and leading to unstable optimization. 
Second, since optimization-based methods optimize output token probabilities, they provide only indirect and weak control over expert selection that governs the model's safety behavior~\cite{chaudhari2025sparsitysuperpositionmixtureexperts, yang2025mixtureexpertsintrinsicallyinterpretable}. Third, optimization-based attacks typically enforce rigid output patterns on the first few tokens. As a result, the model will often initially comply but subsequently refuse or produce incoherent outputs as the generation continues~\cite{qi2024safety, zhu2024advprefix, xie2025beyond}. To our knowledge, there is no optimization-based attack on MoE models under realistic input-only constraints.

\noindent\textbf{Our Solution.}
We present \ourname, a routing-aware adversarial framework that enables effective jailbreak attacks on MoE LLMs under realistic deployment-time input-only constraints. Our key insight is that, although routing decisions are discrete, they are determined by continuous router scores that can be optimized through input perturbations to steer expert selection during inference. Unlike prior output-centric attacks, \ourname directly leverages routing behavior to guide adversarial input construction.

\ourname consists of two components. First, we introduce a \emph{response-driven contrastive profiling method} that identifies safety-critical and harmful experts by comparing activations under paired safe and harmful responses, isolating behavior-relevant experts without being confounded by prompt semantics. Second, we design a \emph{routing-aware optimization method} to construct adversarial suffixes that manipulate expert selection during inference. It jointly suppresses safety experts, promotes harmful experts under a bounded constraint, and discourages early-stage refusal tokens. \ourname aligns optimization with both routing dynamics and autoregressive generation, enabling stable and effective attacks despite non-differentiable routing.
By targeting routing directly, \ourname enables precise control over model behavior through input-only perturbations , avoids brittle output constraints, and does not require any access to model parameters or inference code modifications in the deployment phase. The optimized suffix can be generated offline using a surrogate model and transferred to the target, making the attack practical in real-world settings. Our contributions are as follows:
\begin{itemize}[leftmargin=*]
    \item We propose \ourname, a routing-aware MoE attack framework that enables effective jailbreak attacks under realistic input-only constraints by directly influencing expert selection through optimized inputs.
    \item We introduce a response-driven contrastive profiling method that isolates safety-critical and harmful experts based on generation behavior, providing a precise characterization of the model's internal safety mechanisms.
    \item We design a routing-aware optimization objective that jointly controls expert activation and early-stage decoding dynamics, thereby enabling stable and effective attacks despite routing's non-differentiable nature.
    \item We evaluate \ourname on seven MoE LLMs across three MoE architectures, demonstrating strong attack effectiveness with limited utility loss, achieving an average ASR from 7.1\% to 69.3\%, and only a 1.3\% average drop across five NLU benchmarks.
    \item We show that routing-based vulnerabilities generalize across models and modalities: the optimized suffix raises average ASR from 27.7\% to 61.2\% on five sibling MoE variants and from 2.47\% to 38.7\% on three MoE-based Vision Language Models, highlighting novel security risks in sparse expert architectures.
\end{itemize}

The remainder of this paper is organized as follows. Section~\ref{sec:Preliminaries} provides background on MoE architectures and related attack methods. Section~\ref{sec:threat model} introduces the threat model. Section~\ref{sec:design} presents the \ourname framework. Section~\ref{sec:experimental setup} describes implementation details. 
Section~\ref{sec:Case Study} visualizes the distribution of experts.
Section~\ref{sec:Experimental Results} reports the empirical results. 
Section~\ref{sec:ablation} presents ablation studies, Section~\ref{sec:discussion} discusses the potential defenses, and Section~\ref{sec:conclusion} concludes this work.

\section{Preliminaries}
\label{sec:Preliminaries}
\subsection{Mixture-of-Experts}
\label{sec:Mixture-of-Experts}
The Mixture-of-Experts (MoE) architecture~\cite{jacobs1991adaptive, shazeer2017outrageously} is introduced to address the computational scaling limitations of dense Transformers~\cite{vaswani2017attention}. MoE decouples parameter capacity from inference latency, allowing models to scale in size without a proportional increase in active computational cost (FLOPs).  
Formally, the standard dense Feed-Forward Network (FFN) is replaced by an MoE layer comprising a routing network $G$ and a set of $N$ expert FFNs, $\{e_1, e_2, \ldots, e_N\}$. Given an input $x \in \mathbb{R}^{d}$, the routing network $G$ produces a probability distribution over experts:
\begin{equation}
G(x) = \mathrm{Softmax}(W_g x),
\end{equation}
where $W_g \in \mathbb{R}^{N \times d}$ is the router parameter matrix.
To enforce computational sparsity, only a subset of experts $\mathcal{T} = \mathrm{TopK}(G(x), K)$ with the highest routing scores are activated per token, where $K < N$. The final MoE output is computed as the weighted sum of the selected experts' outputs:
\begin{equation}
\mathrm{MoE}(x) = \sum_{i \in \mathcal{T}} w_i(x)\, e_i(x),
\end{equation}
where $w_i(x)$ denotes the final routing weight assigned to expert $E_i$. Depending on the implementation, $w_i(x)$ may correspond to the raw softmax probability $P_i(x)$ or a re-normalized probability over the selected expert set $\mathcal{T}$, where $\sum_{i \in \mathcal{T}} w_i(x) = 1$.

\noindent\textbf{Architectural Variants.}
The MoE architecture has evolved along several distinct variants. Beyond standard sparse routing (e.g., Mixtral~\cite{mistral2023mixtral}), more recent models such as DeepSeek-MoE~\cite{dai2024deepseekmoeultimateexpertspecialization} introduce shared experts: small subsets of experts that are activated for every token, alongside conditionally routed ones. Other designs, such as Pangu-Pro-MoE~\cite{tang2025pangupro}, adopt grouped routing strategies to balance load across clusters of experts. 
Despite these design differences, sparse activation is the main feature of all MoE variants. Recent mechanistic interpretability research~\cite{chaudhari2025sparsitysuperpositionmixtureexperts, yang2025mixtureexpertsintrinsicallyinterpretable} suggests that sparse routing reduces representation superposition and promotes more monosemantic expert behavior. From a security perspective, this shift has important consequences: safety-aligned behaviors are no longer distributed across the model~\cite{lai2025safexanalyzingvulnerabilitiesmoebased}, but are becoming expert-dependent. This concentration creates a localized and potentially exploitable attack surface across all MoE architectures. In this work, we evaluate all three variants and show that \ourname generalizes across them, highlighting its architecture-agnostic design.

\subsection{Attacks on LLMs}


Safety-aligned LLMs remain vulnerable to adversarial exploitation, ranging from input-level attacks~\cite{dan_jailbreak_prompts} to parameter-level interventions~\cite{qi2024safety,xu2025reasoning,wan2023poisoning}. Among these, input-level attacks are particularly attractive due to its wide applicablity in real-world API-only settings.
Two representative classes of such attacks are vanilla jailbreaks and optimization-based jailbreaks.

\noindent\textbf{Vanilla Jailbreaks.} Early jailbreak approaches rely on manually crafted prompts designed to bypass safety alignment by reframing or obfuscating harmful intent~\cite{liu2023autodan}. Common strategies include role-playing (e.g., asking the model to act as an unrestricted agent), indirect phrasing, or embedding the request within seemingly benign contexts~\cite{yi2024jailbreak}. While often effective, these methods depend heavily on human intuition and extensive trial-and-error, and they tend to be brittle: small changes in wording or model updates can significantly reduce their success rate. Moreover, such prompts typically lack transferability across models and tasks~\cite{lin2025understanding}.

\noindent\textbf{Optimization--based Jailbreaks.}
To overcome these limitations, recent work has shifted toward automated, optimization-based attacks. Greedy Coordinate Gradient (GCG)~\cite{zou2023universal,liu2023autodan,liao2024amplegcg} formulates jailbreak as a discrete optimization problem. Given a harmful query, it appends an adversarial suffix and then updates the suffix tokens over multiple iterations based on the target loss so that a predefined affirmative response (e.g., ``Sure, here is'') becomes more likely. After this search process, the optimized suffix pushes the model toward a compliant answer by exploiting the weakness that safety alignment is often concentrated in the first few tokens~\cite{qi2024safety, zou2023universal}. This objective steers the model toward compliant responses and yields more systematic attacks than manual jailbreaks.

However, conventional GCG attacks face three fundamental limitations when applied to modern MoE architectures. First, GCG is designed for dense models and optimizes a cross-entropy objective at the output layer, requiring gradients to propagate through the full network. In MoE models, this signal is disrupted by the non-differentiable Top-$K$ routing at each layer~\cite{fedus2022switch}, leading to unstable and ineffective suffix optimization. Second, GCG remains output-centric rather than structure-aware: it optimizes token probabilities instead of directly exploiting the routing bottleneck that is unique to MoE models. As a result, its influence over expert selection is only indirect and weak~\cite{cai2024survey}. Third, GCG enforces a rigid affirmative prefix that conflicts with modern alignment mechanisms. Recent work shows that safety alignment is encoded throughout the generation process rather than limited to initial tokens~\cite{qi2024safety}. Consequently, even if the model is forced to begin with an affirmative response, it often reverts to refusal in subsequent tokens (e.g., ``however, I cannot...''). This mismatch between the optimization objective and the model’s internal safety dynamics can destabilize generation, frequently producing incoherent or degenerate outputs instead of successful jailbreaks~\cite{tan2025resurgence}. To the best of our knowledge, there is still no input-level attack that can reliably bypass MoE models while producing coherent, harmful responses.


\subsection{Attacks on MoE}
As MoE architectures become more widely adopted, recent work has begun to examine their distinct security properties. A key observation is that sparse routing introduces attack surfaces that do not arise in dense models. For instance, BadMoE~\cite{wang2025badmoebackdooringmixtureofexpertsllms} shows that adversaries can implant backdoors into dormant experts: experts that are rarely activated during training, without degrading overall model performance. Post-training alignment in MoE models is also structurally fragile. SAFEx~\cite{lai2025safexanalyzingvulnerabilitiesmoebased} and SteerMoE~\cite{fayyaz2026steeringmoellmsexpert} find that refusal behavior is concentrated in a small subset of experts, often referred to as safety experts. By identifying these experts through activation profiling on harmful prompts, they demonstrate that disabling them at inference time can reliably bypass safety mechanisms. Similar ideas appear in L$^3$~\cite{lintelo2026large}, which uses sequence modeling to identify critical experts, and F-SOUR~\cite{jiang2026sparse}, which perturbs routing through randomized search. Despite exposing this structural weakness, these methods are not input-only: they require direct control over the model's forward pass, e.g., through expert masking, pruning, or logit manipulation. This requirement is unrealistic in typical deployment settings, where attackers can interact with the model only through input prompts.

Our work addresses this gap by focusing on routing manipulation. In the offline stage, we use access to an open-weight MoE model to profile safety-related experts and optimize an adversarial suffix that changes expert activation patterns. At deployment time, the attack is purely input-level: the adversary only appends the learned suffix to the prompt. This suffix shifts routing away from safety experts and toward harmful pathways, giving a similar effect to expert-level interventions without modifying model parameters, inference code, or decoding at inference time.

\section{Threat Model}
\label{sec:threat model}
We consider transfer attacks across intra-family MoE variants. This setting reflects common LLM deployment practice, where open-weight backbones are often repackaged into proprietary assistants, managed APIs, and domain-specific products. Reports from McKinsey and IBM are consistent with this trend: 63\% of organizations use open-source AI models, and 60\% obtain AI tools from open-source ecosystems~\cite{bisht2025opensourceai,ibm2024opensourceai}. Platforms such as OpenRouter, which aggregates 300+ models from 60+ providers, further illustrate such reuse~\cite{openrouter2026about}. Meanwhile, due to the scaling and performance advantages of MoE, more recently released models adopt MoE architectures~\cite{mistral2023mixtral,dai2024deepseekmoeultimateexpertspecialization,qwen_moe_2024,yang2025qwen3,tang2025pangupro}. Routing-level vulnerabilities may therefore persist across related downstream variants within the same family.

\noindent
\textbf{Adversary Capabilities.}
We assume a two-stage, proxy-based transfer setting. The adversary performs offline optimization on a surrogate model to attack a target model at deployment time.
\begin{itemize} [leftmargin = *]
    \item Offline Stage: The adversary has full white-box access to a closely related open-weight MoE model (e.g., a public checkpoint from the same family as the target). This includes access to weights, intermediate activations, and router logits, which are used to localize experts and optimize a universal adversarial suffix.
    \item Deployment Stage: The attack is strictly input-level. Unlike invasive methods (e.g., SteerMoE, SAFEx, or L$^3$), the adversary cannot modify weights, prune experts, or alter the inference code of the target system. The adversary can only submit text prompts containing a discrete adversarial suffix.
\end{itemize}

\noindent
\textbf{Adversarial Objective.}
The adversary’s goal is to find a universal adversarial suffix that, when appended to a harmful query, consistently steers routing away from ``safety-aligned'' experts and toward pathways associated with harmful content. A successful attack must bypass safety alignment while maintaining the model’s fluency and general reasoning capabilities. 

\input{img/img_main}

\section{\ourname}
\label{sec:design}
\ourname is a routing-aware adversarial framework for MoE models. An overview of the framework is shown in Figure~\ref{fig:framework}. We first localize safety- and harm-related experts via response-driven contrastive profiling that disentangles behavior from prompt semantics. We then optimize a universal adversarial suffix using a novel ternary loss that directly steers expert routing at inference time. Together, these components enable precise and effective manipulation of MoE behavior through input-level suffix injection. Once optimized offline on an open-weight source model, the suffix can be transferred to related MoE variants without modifying the target model's parameters, inference code, or decoding strategy.

\subsection{Expert Localization}
\label{section:Expert Localization}

Prior work typically identifies critical experts by comparing activation frequencies under malicious and benign prompts~\cite{lai2025safexanalyzingvulnerabilitiesmoebased, wang2025badmoebackdooringmixtureofexpertsllms, wu2025gatebreaker, lintelo2026large, wu2025neurostrike}. However, this prompt-centric strategy is inherently coarse-grained: differences in expert activation are entangled with prompt semantics (e.g., topic, style, or domain), rather than isolating safety-relevant behavior. As a result, experts who respond to content variation (e.g., ``how to make a cake'' vs.\ ``write a spam email'') can be misattributed to safety mechanisms, introducing substantial noise. To obtain a more precise decomposition, we shift the analysis from prompts to model responses. Specifically, we perform \emph{response-driven contrastive profiling}, where a malicious query $x_{\mathrm{query}}$ is paired with two responses: a safe refusal $a_{\mathrm{safe}}$ and a compliant harmful answer $a_{\mathrm{harm}}$. Using teacher forcing, we run forward passes on $(x_{\mathrm{query}} \oplus a_{\mathrm{safe}})$ and $(x_{\mathrm{query}} \oplus a_{\mathrm{harm}})$, masking the query tokens and collecting routing statistics only over the generated responses. This isolates expert behavior conditioned on the same input, eliminating confounding factors from prompt semantics.

We quantify expert behavior using normalized activation frequencies. Let $F_l(e \mid a)$ denote the frequency of expert $e$ at layer $l$ when processing answer $a$:
\begin{equation}
F_l(e \mid a) = \frac{1}{|a|} \sum_{t=1}^{|a|} \mathbb{I}\left(e \in \mathcal{A}_{l,t}\right),
\label{eq:frequency of expert}
\end{equation}
where $\mathcal{A}_{l,t}$ is the set of activated experts at response token $t$, and $|a|$ denotes the number of valid response tokens in answer $a$ after masking the query tokens. We then define the \emph{safety differential}:
\begin{equation}
\Delta_S(l,e) = F_l\!\left(e \mid a_{\mathrm{safe}}\right) - F_l\!\left(e \mid a_{\mathrm{harm}}\right),
\label{eq:safety differential}
\end{equation}
which measures how strongly an expert is associated with refusal versus harmful generation. Experts with large positive $\Delta_S(l,e)$ are strongly tied to safety behavior, while those with large negative values are more active in harmful responses. We visualize the safety differential in Section~\ref{sec:Case Study} .

\subsubsection{Utility-Aware Filtering}
\label{section:Utility Penalization}
While Eq.~\ref{eq:safety differential} reliably distinguishes the safety contribution of each expert, it poses a critical challenge on polysemanticity: some experts may contribute to both safety enforcement and general language modeling. Directly manipulating such experts risks degrading overall model utility. To address this, we introduce utility-aware filtering. Specifically, using a benign instruction-following dataset $\mathcal{D}_{\mathrm{gen}}$, we estimate the general activation frequency $P_l(e \mid \mathcal{D}_{\mathrm{gen}})$ and define:
\begin{equation}
\begin{cases}
\mathrm{Score}_{\mathrm{safe}}(l,e) = \Delta_S(l,e) - \left(P_l\!\left(e \mid \mathcal{D}_{\mathrm{gen}}\right)\right)^2, \\
\mathrm{Score}_{\mathrm{harm}}(l,e) = \Delta_S(l,e).
\end{cases}
\end{equation}

The quadratic penalty suppresses high-frequency general-purpose experts when selecting safety targets, ensuring that the selected safety experts are behaviorally specialized. In contrast, we do not penalize harmful experts: retaining experts with strong general capabilities improves both optimization stability and generation fluency during attack construction.
Finally, we globally rank all layer-expert pairs $(l,e)$ by their scores. We select the top-$N$ pairs with the highest $\mathrm{Score}_{\mathrm{safe}}(l,e)$ to form the safety target set $\mathcal{E}_{\mathrm{safe}}$, and the top-$N$ pairs with the lowest $\mathrm{Score}_{\mathrm{harm}}(l,e)$ to form the harmful target set $\mathcal{E}_{\mathrm{harm}}$. Here, each element of $\mathcal{E}_{\mathrm{safe}}$ or $\mathcal{E}_{\mathrm{harm}}$ is a selected layer-expert pair. For convenience, let $L_{\mathrm{safe}} = \{\, l \mid \exists e,\ (l,e) \in \mathcal{E}_{\mathrm{safe}} \,\}$ and $L_{\mathrm{harm}} = \{\, l \mid \exists e,\ (l,e) \in \mathcal{E}_{\mathrm{harm}} \,\}$ be the layers that appear in the selected pairs. For each layer $l$, we define $\mathcal{E}_{\mathrm{safe}}^{(l)} = \{\, e \mid (l,e) \in \mathcal{E}_{\mathrm{safe}} \,\}$ and $\mathcal{E}_{\mathrm{harm}}^{(l)} = \{\, e \mid (l,e) \in \mathcal{E}_{\mathrm{harm}} \,\}$. To validate the design of our utility-aware filtering, we provide the ablation study in Section~\ref{section:Impact of Utility Penalization}.

\subsection{Ternary-Loss for Route Hijacking}
\label{sec:Ternary-Loss for Routing Hijacking}
Given the selected safety pair set $\mathcal{E}_{\mathrm{safe}}$ and harmful pair set $\mathcal{E}_{\mathrm{harm}}$, our goal is to construct a universal adversarial suffix $x_{\mathrm{adv}}$ that systematically manipulates MoE routing at inference time. For a malicious query $x_{\mathrm{query}}$, we optimize over the combined input $x_{payload} = x_{\mathrm{query}} \oplus x_{\mathrm{adv}}$ and denote by $t^\star = |x_{payload}|$ the final prompt token, i.e., the last input token before autoregressive decoding begins. We directly target the pre-truncation router distribution $p_{l,e}^{(t^\star)}(x_{payload})$ at this boundary token and the selected layer-expert locations. Unlike prior methods such as GCG, which operate on output token probabilities and thus only indirectly influence model behavior, our approach intervenes at the routing level, the primary control mechanism in MoE models. Since the router determines which experts are activated, and expert specialization governs whether the model follows safe or harmful computation paths, routing-level optimization provides a more direct, efficient, and architecture-aligned means of steering model behavior.

Our route hijacking objective is grounded in three structural properties of MoE models: safety behavior is localized in a small subset of experts~\cite{lai2025safexanalyzingvulnerabilitiesmoebased, lintelo2026large}; harmful generation arises from alternative expert pathways that are suppressed during alignment~\cite{fayyaz2026steeringmoellmsexpert}; and refusal behavior is reinforced during autoregressive decoding through characteristic token patterns~\cite{qi2024safety, zou2023universal}. 
These observations imply that a successful attack must simultaneously (1) suppress safety experts without disrupting the model's general capability, (2) activate harmful experts and (3) prevent the decoder from reverting to refusal trajectories. This motivates our ternary-loss formulation.

\noindent\textbf{(1) Safety Suppression ($\mathcal{L}_{\mathrm{suppress}}$).}
The selected safety experts $\mathcal{E}_{\mathrm{safe}}^{(l)}$ are explicitly chosen to be highly specialized for refusal behavior. Their activation causally increases the probability of safety refusal. Therefore, the most direct way to disable safety is to reduce their routing probability:
\begin{equation}
\mathcal{L}_{\mathrm{suppress}} = \frac{1}{|L_{\mathrm{safe}}|} \sum_{l \in L_{\mathrm{safe}}} \sum_{e \in \mathcal{E}_{\mathrm{safe}}^{(l)}} p_{l,e}^{(t^\star)}(x_{payload}),
\end{equation}
where $L_{\mathrm{safe}}$ is the set of layers that appear in $\mathcal{E}_{\mathrm{safe}}$, and $\mathcal{E}_{\mathrm{safe}}^{(l)}$ is the set of selected safety experts in layer $l$. Minimizing $\mathcal{L}_{\mathrm{suppress}}$ reduces the chance that these safety experts are selected by the Top-$K$ router at the boundary token $t^\star$, effectively weakening the model's primary defense mechanism before generation begins.

\noindent\textbf{(2) Bounded Harmful Promotion ($\mathcal{L}_{\mathrm{promote}}$).}
Suppressing safety experts alone does not guarantee harmful generation, as the router may still favor neutral experts. We must actively increase the activation of the selected harmful experts $\mathcal{E}_{\mathrm{harm}}^{(l)}$. 
However, naively maximizing their probability leads to degenerate routing (i.e., collapsing onto a few experts), which harms fluency and reduces transferability. Instead, we impose a \emph{bounded activation constraint}:
\begin{equation}
\mathcal{L}_{\mathrm{promote}} = \frac{1}{|L_{\mathrm{harm}}|} \sum_{l \in L_{\mathrm{harm}}} 
\max\left(0, m_{\mathrm{harm}} - \sum_{e \in \mathcal{E}_{\mathrm{harm}}^{(l)}} p_{l,e}^{(t^\star)}(x_{payload})\right).
\end{equation}
This hinge loss enforces that, at the boundary token $t^\star$ and in each selected harmful layer, the total routing mass on the selected harmful experts exceeds a threshold $m_{\mathrm{harm}}$. Importantly, this design aligns with the Top-$K$ routing mechanism: increasing the cumulative probability makes the promoted harmful experts more likely to be selected by the router, and once the threshold is reached, further maximization becomes unnecessary. This prevents over-optimization and preserves the contribution of general-purpose experts, ensuring fluent generation.

\noindent\textbf{(3) Refusal Unlikelihood ($\mathcal{L}_{\mathrm{refusal}}$).} 
Even when routing is successfully manipulated, modern aligned models can still revert to refusal due to learned decoding priors (e.g., preferring tokens like ``I'', ``cannot'', ``sorry'' at early steps). Prior works~\cite{qi2024safety, zou2023universal} show that such early tokens strongly determine the generation trajectory. To counter this, we introduce a token-level unlikelihood objective. We construct a refusal vocabulary $\mathcal{V}_{\mathrm{refuse}}$ from refusal templates aggregated from the refusal responses produced by the evaluated models on harmful queries; the full template list is given in Appendix~\ref{appendix:refusalTemplates}. We then decompose these templates into subword tokens and penalize their generation over the first $W$ decoding steps. We empirically set the window size $W=5$, as safety guardrails typically trigger refusal behaviors within these initial tokens:
\begin{equation}
\mathcal{L}_{\mathrm{refusal}} = -\frac{1}{W} \sum_{t=1}^{W} 
\log\left(1 - \sum_{y \in \mathcal{V}_{\mathrm{refuse}}} 
p_\theta\left(y \mid x_{payload}, y_{<t}\right)\right),
\end{equation}
where $p_\theta(\cdot)$ represents the model's output probability distribution, and $y_{<t}$ denotes the sequence of tokens generated before the $t$-th token. 
This formulation has two advantages: (i) it directly suppresses early refusal triggers, which are causally important for alignment, and (ii) it avoids the combinatorial complexity of sequence-level objectives, providing stable gradients for discrete optimization.

\input{tab/tab_targetModels}

\noindent\textbf{Final Loss for Routing Hijacking.} The final loss is defined as:
\begin{equation}
\mathcal{L}_{\mathrm{total}} = \lambda_1 \mathcal{L}_{\mathrm{suppress}} + \lambda_2 \mathcal{L}_{\mathrm{promote}} + \lambda_3 \mathcal{L}_{\mathrm{refusal}},
\end{equation}
where $\lambda_1, \lambda_2, \lambda_3$ balance routing suppression, harmful activation, and decoding control. Indeed, the ternary formulation reflects three distinct failure modes in MoE alignment. Suppressing safety experts alone is insufficient without redirecting routing toward harmful experts, while routing manipulation alone does not prevent the decoder from reverting to refusal behavior. Each component, therefore, targets a necessary stage of the MoE computation pipeline, i.e., expert selection and autoregressive generation. Ablation in Section~\ref{sec:Ablation of Ternary-Loss Components} confirm that removing any term degrades attack success.

\subsection{Optimization Pipeline}
\label{subsec:Optimization Pipeline}
Optimizing the discrete adversarial suffix $x_{\mathrm{adv}}$ under routing-based objectives is fundamentally challenging due to both discrete token representations and the hard Top-$K$ routing in MoE layers. We build on gradient-guided discrete optimization~\cite{zou2023universal}, but shift the optimization target from output token likelihoods to \emph{routing-level control}. Unlike GCG, which indirectly steers model behavior through the final language modeling head, our method directly manipulates the router, thereby intervening at the point where safety behavioral decisions are made.

The full optimization pipeline is summarized in Algorithm~\ref{alg:optimization}. At each iteration, we compute gradients of $\mathcal{L}_{\mathrm{total}}$ with respect to the one-hot encoding of $x_{\mathrm{adv}}$, backpropagating through the continuous router softmax probabilities at the boundary token $t^\star$ prior to Top-$K$ truncation. This provides a differentiable surrogate for otherwise discrete routing decisions, enabling gradient-based optimization over expert selection.
Next, we construct candidate updates by selecting tokens with the largest negative gradients at each position, approximating steepest descent directions. A key technical challenge arises from subword tokenization: token substitutions can alter sequence segmentation after decoding and re-encoding, leading to misalignment between token positions and routing states. To address this, we enforce a strict decode-then-re-encode constraint and discard candidates whose tokenized length deviates from $T$.
Finally, we evaluate all valid candidates in batch and update the suffix by selecting the one that minimizes $\mathcal{L}_{\mathrm{total}}$. Indeed, this pipeline enables stable and effective routing manipulation by (i) bypassing the non-differentiable Top-$K$ gating via soft routing and (ii) preserving token-level alignment required for consistent expert selection.

\section{Implementations Details}
\label{sec:experimental setup}

\subsection{Expert Profiling}
\label{sec:Expert Profiling}
To identify safety and harmful experts via response-driven contrastive profiling (as introduced in Section~\ref{section:Expert Localization}), we sample 600 paired harmful and safe completions from the LLM-LAT dataset~\cite{sheshadri2024latent}, which has 4,950 paired harmful completions and safe refusal completions for identical malicious prompts. Concurrently, to calculate the utility penalty and filter out polysemantic general-purpose experts, we construct a benign corpus by randomly sampling 600 instances from a mixture of the Alpaca~\cite{peng2023instruction} and WikiText-2~\cite{merity2016pointer} datasets. During the forward-pass profiling across these corpora, for both contrastive profiling and utility filtering, we mask all input query tokens and chat-template special tokens, ensuring that activations are collected exclusively on the generated response tokens. Finally, to eliminate sequence-length bias between lengthy harmful completions and concise refusals, the raw expert activation counts are normalized by response token number, yielding the normalized frequency $F_l(e \mid a)$ defined in Eq.~\eqref{eq:frequency of expert}.

\input{tab/tab_benchmarking}

\subsection{Ternary-loss Hyperparameters}
\label{sec:Optimization Hyperparameters}
\input{algorithm/alg}
The overall optimization behavior is controlled by the weighting coefficients of three loss components: safety suppression ($\lambda_1$), harmful promotion ($\lambda_2$), and refusal unlikelihood ($\lambda_3$). To determine the optimal configuration, we conducted a preliminary grid search over the ranges $\lambda_1 \in [1, 4]$ and $\lambda_2, \lambda_3 \in [1, 2]$. Based on this search, we set the coefficients to a ratio of 3:1:1 ($\lambda_1=3, \lambda_2=1, \lambda_3=1$), which provides the balance between suppressing safety behavior and preserving generation fluency. We then used the same grid-search results to choose the boundary parameter for harmful expert promotion. For the targeted promotion of harmful experts ($\mathcal{L}_{\mathrm{promote}}$), we use a bounded loss that only pushes the cumulative routing mass on the selected harmful experts until it reaches a margin threshold $m_{\mathrm{harm}} = 0.3$. To determine this value, we further examined runs whose outputs degenerated into gibberish and measured the summed activation frequency of the promoted harmful experts. We found that when this quantity was pushed much beyond 0.3, a small set of harmful experts began to dominate the generation process, which hurt output fluency. Setting $m_{\mathrm{harm}} = 0.3$ therefore gives a practical balance: it keeps the attack effective while avoiding expert monopolization and preserving fluent outputs. Section~\ref{sec:Ablation of Ternary-Loss Components} discusses the rationale behind the hyperparameter setting.

\subsection{Optimization Configuration}
\label{sec:Optimization Configuration}
Following Algorithm~\ref{alg:optimization}, we adopt a universal attack setting where a single adversarial suffix is optimized across 16 malicious prompts and then applied at deployment time~\cite{zou2023universal}. We treat the individual, per-prompt optimization mode only as an upper-bound ablation and discuss it in Section~\ref{abl:Suffix Settings}. After expert localization, we construct the safety and harmful target sets by selecting the top 20\% of ranked layer-expert pairs for $\mathcal{E}_{\mathrm{safe}}$ and $\mathcal{E}_{\mathrm{harm}}$, respectively; the sensitivity to this proportion is analyzed in Section~\ref{sec:expert-proportion}. Following prior works~\cite{zou2023universal, liao2024amplegcg}, at each step, we construct a candidate batch of size 128 by selecting replacement tokens from the top $k=256$ negative gradient indices, we set the adversarial suffix length to $T=20$ and use 300 optimization steps. This relatively small optimization budget is sufficient because, although the ternary loss includes an output-level refusal penalty, the search is still primarily guided by gradients on the localized routing probabilities, making \ourname substantially more efficient than standard GCG attacks. A detailed discussion of the GCG configuration and the corresponding efficiency comparison is provided in Appendix~\ref{appendix:gcgAblation}.

\subsection{Evaluation Metrics}
\label{section:Evaluation Metrics}

\noindent\textbf{Attack Success Rate (ASR).} We evaluate the effectiveness of the adversarial suffix by measuring ASR: the percentage of malicious prompts that trigger toxic or policy-violating responses. For robust assessment, we use a multi-stage pipeline: \textit{Llama-Guard-3-8B}~\cite{dubey2024llama3herdmodels} as the primary safety judge, with \textit{Qwen3Guard-Gen-8B}~\cite{zhao2025qwen3guard} and human verification to resolve ambiguous cases and false positives.
    
\noindent\textbf{General Utility.} We evaluate the general utility of the target models before and after appending the adversarial suffix via CoLA (linguistic acceptability)~\cite{warstadt2019neural}, RTE (inference)~\cite{dagan2005pascal}, WinoGrande (commonsense reasoning)~\cite{sakaguchi2021winogrande}, OpenBookQA (general knowledge)~\cite{mihaylov2018can}, and ARC-Challenge (grade-school science)~\cite{clark2018think}.

\noindent\textbf{Mechanistic Validation of Routing Hijacking.} To assess whether routing shifts drive ASR, we evaluate two aspects: \textit{Boundary Shift} and \textit{Global Shift}.
In Boundary Shift, we define \textit{Target Expert Suppression Rate (TESR)} and \textit{Target Harmful Promotion Rate (THPR)} at the last input token $t^\star$, measuring changes in pre-truncation routing mass. TESR captures the reduction for safety experts ($\mathcal{E}_{\mathrm{safe}}$), while THPR measures the increase for harmful experts ($\mathcal{E}_{\mathrm{harm}}$).
In Global Shift, we track Top-$K$ activation frequencies of target experts across all generated tokens to verify the persistence of routing shifts.

\section{Case Study: Visualize Safety Experts}
\label{sec:Case Study}
The ability to localize safety-critical experts is fundamental to the success of \ourname. To investigate how safety alignment is distributed across a model’s architecture, we analyze the sparse expert behavior of DeepSeek-MoE-A2.7B-Chat. Specifically, for each layer-expert pair $(l,e)$, we visualize the safety differential $\Delta_S(l,e)$ (see Eq.~\eqref{eq:safety differential}), which measures the difference in activation frequency between safe and harmful completions for identical malicious queries.

\begin{figure}[t]
    \centering
    \includegraphics[width=1\linewidth]{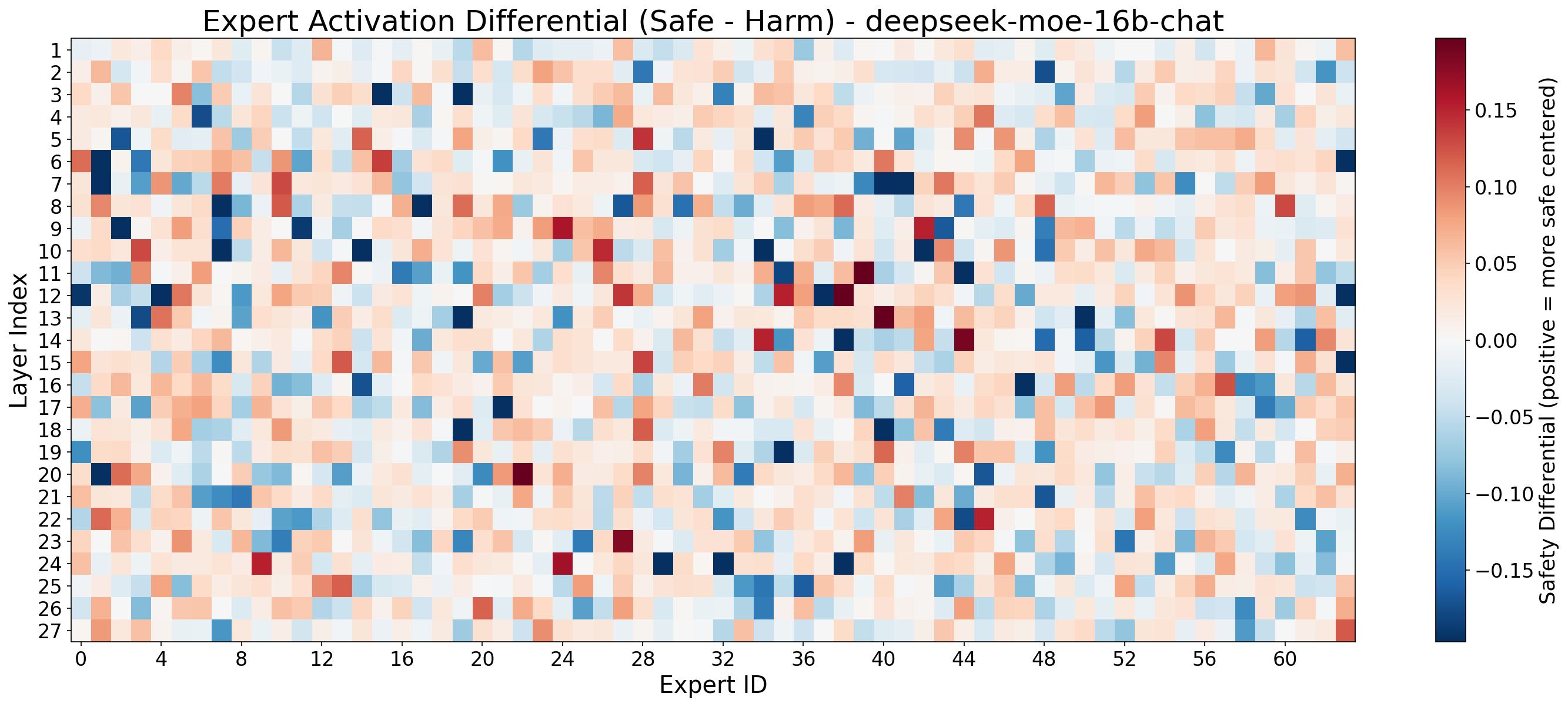}
    \caption{Safety differential heatmap for DeepSeek. X- and Y-axis represent expert ID and layer index, respectively.}
    \label{fig:heatmapDeepseek}
\end{figure}

As shown in Figure \ref{fig:heatmapDeepseek}, the model's safety behavior is highly sparse and localized. While most experts remain neutral (near zero), a distinct subset shows significant activation bias (e.g., $\lvert \Delta_S \rvert > 0.15$). More specifically, two types of experts can be identified.
\begin{itemize} [leftmargin = *]
    \item Safety Experts (Warm regions): Clusters with high positive differentials indicate experts that are preferentially activated when the model refuses a harmful request.
    \item Harmful Experts (Cool regions): Areas with high negative differentials highlight experts who are more active during the generation of harmful content.
\end{itemize}

The coexistence of these specialized regions confirms a functional separation within the MoE backbone: safety alignment is not uniformly distributed but is instead concentrated within specific "safety-aligned" experts. This structural locality provides the leverage necessary for \ourname to compromise the model via router-level manipulation. Similar sparsity patterns across other evaluated MoE models are provided in Appendix \ref{appendix:Visualization of Safety Differential}.



\input{tab/tab_generalAbility}

\section{Experimental Results}
\label{sec:Experimental Results}

Table~\ref{tab:targetModels} summarizes the diverse set of state-of-the-art MoE LLMs evaluated in our experiments. To ensure a comprehensive evaluation, our target models encompass the three distinct routing strategies discussed in Section~\ref{sec:Mixture-of-Experts}, i.e., standard sparse routing, shared-expert mixtures, and grouped mixtures, alongside varying reasoning paradigms (CoT and non-CoT), with total model sizes ranging broadly from 14.3B to 80.4B parameters which represent the latest and mostly adopted MoE LLMs.

We benchmark \ourname against three adversarial baselines. For input-level attacks, we compare with GCG~\cite{zou2023universal}, using the same setup as in Section~\ref{sec:Optimization Configuration} with a 500-step optimization budget (see Appendix~\ref{appendix:gcgAblation}) and targeting the affirmative prefix ``Sure, here is''. For white-box inference-time attacks, we include SAFEx~\cite{lai2025safexanalyzingvulnerabilitiesmoebased} and SteerMoE~\cite{fayyaz2026steeringmoellmsexpert} as two most-recent MoE-specific attacks that directly intervene on expert behavior during inference, through expert pruning and logit steering, respectively. Following their original protocols, for SAFEx, we profile expert activations over 20,000 jailbreak prompts and prune safety-critical experts at inference, while for SteerMoE, we perturb expert logits before top-$K$ routing to steer model behavior. Attack performance is evaluated on the StrongREJECT benchmark~\cite{souly2024strongreject}, which contains 313 malicious prompts spanning several categories.

\subsection{Attack Efficacy and Benchmarking}

Table~\ref{tab:benchmarking} presents the ASR benchmark for the baseline (directly evaluated on malicious prompts), GCG, SAFEx, SteerMoE, and \ourname. Our results show that \ourname outperforms all counterparts, achieving an average ASR of 69.3\%. This success rate is more than double that of the best-performing white-box baseline, SAFEx (33.3\%), and more than triple that of standard GCG (21.9\%). Notably, this effectiveness is consistent across all three major MoE architectures. This indicates that the localized nature of safety alignment represents a universal architectural vulnerability in MoEs, making them susceptible to \ourname regardless of their specific routing design. Interestingly, MoE models with a large number of experts, such as DeepSeek-MoE-16B and Qwen1.5-MoE-A2.7B, are highly susceptible to our attack, reaching ASRs near 90\%. This high success rate occurs because their strong expert specialization allows \ourname to efficiently isolate and bypass safety pathways. In contrast, models with coarse-grained routing (i.e., fewer experts per layer), such as Phi-3.5-MoE (16 experts in each layer), exhibit greater resistance. Indeed, with fewer experts available, safety mechanisms and general linguistic capabilities become heavily entangled, reducing the likelihood that routing manipulation can bypass safety without degrading text generation. Mixtral-8x7B-Instruct-v0.1 is an exception among such coarse-grained models; despite having only 8 experts, it shows a high ASR of 75.1\%. This vulnerability is likely due to weaker initial safety alignment, as indicated by its baseline ASR (14.6\%). Simply shifting the routing distribution away from primary refusal pathways is sufficient to elicit harmful responses.

To further elaborate on the performance gap between \ourname and existing methods, we analyzed the failure cases of the existing methods and identified two primary causes. First, white-box interventions, i.e., SAFEx and SteerMoE, rely on hard pruning or directly manipulating expert logits, which disrupts the internal balance of the MoE layer, frequently leading to semantic collapse and the generation of linguistic gibberish. Second, standard input-level attack GCG fails for two distinct reasons. On one hand, forcing a complex MoE model to output a predefined affirmative response (e.g., ``Sure, here is'') causes severe gradient conflicts~\cite{li2025exploiting, wang2024attngcg}, which destabilizes the routing distribution and produces nonsensical outputs. On the other hand, even when GCG successfully forces the model to generate the initial affirmative tokens, the model's inherent safety alignment often causes it to immediately revert to a refusal in subsequent tokens (e.g., ``Sure, here is how to build a bomb, however, I cannot...'')~\cite{qi2024safety}. \ourname effectively addresses these limitations through its carefully designed ternary-loss objective. Instead of dictating a rigid textual anchor, our refusal unlikelihood penalty ($\mathcal{L}_{\mathrm{refusal}}$) ensures the model does not generate refusal tokens during the first few decoding steps, preserving generation flexibility. Furthermore, rather than directly pruning experts, our method softly manipulates the routing probabilities by suppressing safety experts and promoting harmful ones. This approach guides the model toward compliant pathways without structurally breaking the MoE layers, ensuring continuous and coherent generation while successfully bypassing safety guardrails. We provide attack examples with the optimized suffix in Appendix~\ref{appendix:attackExample}.

\noindent\textbf{General Utility.}
An effective adversarial attack must bypass safety alignments without compromising the model's general linguistic and cognitive capabilities. The results in Table~\ref{tab:generalAbility} confirm that \ourname preserves model utility. Specifically, for each benchmark example, we append the optimized adversarial suffix to the original evaluation prompt and compare the task performance before and after adding the suffix. Across all seven models, the attack incurs minimal degradation in general language abilities. On average, the absolute performance drop is 0.3\% for WinoGrande, 0.2\% for ARC-Challenge, and less than 2.5\% across the syntax (CoLA) and inference (RTE) benchmarks. This minimal utility loss demonstrates the effectiveness of the utility-aware filtering introduced in Section~\ref{section:Utility Penalization}. By excluding polysemantic experts (those responsible for both safety monitoring and foundational syntactic or logic structures) from the target set, our optimization framework focuses on suppressing specialized safety experts without disrupting the model's general-purpose generation pathways.

\subsection{Routing Shift Analysis}
\label{sec:routing_shift}

To verify that \ourname's high ASR stems from router manipulation rather than superficial prompt distraction, we quantify the internal routing shifts. As defined in Section~\ref{section:Evaluation Metrics}, we evaluate two perspectives: the Boundary Shift and the Global Shift.

\input{tab/tab_routerShift}

\noindent\textbf{Boundary Shift.}
As shown in Table~\ref{tab:routerShift}, across all models, \ourname achieves an average Target Expert Suppression Rate (TESR) of -28.75\% and a Target Harmful Promotion Rate (THPR) of +28.51\%. Shifting nearly 30\% of the probability mass in the pre-truncation softmax space demonstrates significant control over the model's expert selection. This shift is particularly obvious in fine-grained MoE architectures (i.e., large expert per-layer), such as Qwen3-30B-A3B-Instruct, where safety expert activation decreases by 65.70\% while dormant harmful expert activation increases by 49.75\%. Similar results in Pangu-Pro-MoE-72B and Qwen1.5-MoE-A2.7B-Chat demonstrate that structurally isolated safety mechanisms are highly vulnerable to targeted routing manipulation.

\noindent\textbf{Global Shift.} Although the routing objectives of our ternary-loss, i.e., $\mathcal{L}_{\mathrm{suppress}}$ and $\mathcal{L}_{\mathrm{promote}}$, are applied only at $t^\star$, routing manipulation exhibits a propagating effect during generation~\cite{queipo2025attention,xu2025steering,chen2025persona}. Table~\ref{tab:routerShift} confirms that the initial context injection influences the entire autoregressive decoding trajectory: global safety expert activation decreases by 23.44\% on average, while harmful expert activation increases by 22.87\%. Mechanistically, our refusal unlikelihood penalty ($\mathcal{L}_{\mathrm{refusal}}$) prevents the generation of standard refusal templates. Constrained by this penalty and guided by the routing manipulation, the model is steered toward generating a compliant response. During autoregressive decoding, the key and value tensors computed from these early generated tokens are cached and attended to by later positions. This updated context changes subsequent hidden states and router logits, allowing the routing shift to persist beyond the boundary token and continue promoting harmful generation.

While most models exhibit synchronized boundary and global shifts, Hunyuan-A13B-Instruct presents an exception: its boundary safety probability slightly increased (+4.64\%), suggesting resistance to initial suppression. Nevertheless, the model ultimately exhibited a global safety drop (-6.90\%) and a harmful promotion (+10.64\%). We attribute this to a delayed routing shift. Although the initial routing was not fully shifted away from safety experts, $\mathcal{L}_{\mathrm{refusal}}$ prevented the model from generating an immediate refusal and instead encouraged a compliant preamble. Once this compliant prefix was established, subsequent decoding steps progressively shifted router logits away from safety experts and toward harmful experts. This demonstrates that \ourname bypasses safety mechanisms even when models show initial resistance to routing manipulation.

\input{tab/tab_transfer}

\subsection{Transferability Across Sibling Models}
A key practical question is whether a suffix optimized offline on one model transfers to other models within the same MoE family. We therefore study zero-shot transfer within each MoE family: for each family, we optimize a universal adversarial suffix on a source model and directly apply it to a target model from the same family without further updates.

As shown in Table~\ref{tab:transfer}, Qwen3-30B-A3B-Thinking-2507 exhibits a low baseline ASR (0.6\%) due to its stronger CoT alignment. However, the suffix optimized on its non-thinking counterpart still raises the ASR to 27.8\% without any target-specific tuning. Similarly, transferring the suffix from Mixtral-8x7B-Instruct to notux-8x7b-v1, a sibling variant fine-tuned for human preference via DPO, increases the ASR from 23.9\% to 41.7\%. These results suggest that, while post-training alignment may modify surface behavior, the underlying routing scheme is largely preserved, enabling transfer attacks with \ourname. The same pattern appears across domain-adapted variants. When transferring the suffix from the general-purpose Qwen1.5-Chat model to the Wikihow variant, the ASR increases from 18.5\% to 83.3\%. We also observe a high baseline ASR of 88.9\% on Qwen1.5-MOE-sft-nemotron-code, indicating that this code-specialized variant is already weakly aligned on our benchmark; even so, \ourname further raises the ASR to 95.9\%. Overall, the average ASR rises from 27.7\% to 61.2\%, demonstrating the strong transferability of \ourname. Beyond zero-shot transfer, we also examine whether the offline optimization process can be learned as a lightweight adversarial suffix generator that maps malicious prompts directly to suffixes. The training setup and preliminary results of this generative variant are provided in Appendix~\ref{sec:adversarial suffix generator}.

\subsection{Transferability Across Modalities}
\label{sec:vlm_attack}

\ourname works beyond text-to-text LLMs. To demonstrate that the routing vulnerabilities exposed by \ourname are inherent to the MoE architecture, we evaluate \ourname against MoE-based Vision Language Models (VLMs). We select three targets: the recently released Qwen3.5 series (35B and 122B parameter variants)~\cite{qwen3.5_2026} and Kimi-VL-A3B-Instruct~\cite{team2025kimi}. These architectures employ a mixture-of-experts language backbone augmented with a vision encoder to handle image inputs~\cite{bordes2024introduction}. We first apply \ourname pipeline on the text modality to generate the adversarial suffix. Once finished, we concatenate it to the malicious prompt and render the entire sequence into a single image, following prior works~\cite{gong2025figstep,liu2025survey}.

\input{tab/tab_visionAttack}

As shown in Table~\ref{tab:visionAttack}, \ourname generalizes well to all VLMs, increasing the average ASR from 2.47\% to 38.7\%. Notably, the Qwen3.5 family exhibited a 0.0\% baseline ASR. However, appending our text-optimized adversarial suffix successfully manipulated the routing of the visual embeddings within the MoE backbone, raising the ASR to over 31\%. These results confirm a significant architectural vulnerability: multimodal safety alignment in VLMs is heavily dependent on the underlying MoE language component. Since visual tokens and text tokens share the same routing subspace, manipulating the router's state with our text-derived suffix bypasses the safety mechanisms, causing the MoE experts to process and comply with the malicious visual intent.

\section{Ablation and Hyperparameter Study}
\label{sec:ablation}


\subsection{Prompt- vs. Response-driven Profiling}
\label{sec:Response-Driven vs. Prompt-Driven}

As detailed in Section~\ref{section:Expert Localization}, \ourname employs response-driven contrastive profiling to isolate functional safety experts, diverging from conventional prompt-centric methods~\cite{lai2025safexanalyzingvulnerabilitiesmoebased, wu2025gatebreaker, wang2025badmoebackdooringmixtureofexpertsllms}. To evaluate this design choice, we identify the top 20\% of safety experts using prompt-driven profiling (collecting activations based on input prompts) and our response-driven approach, respectively, then apply \ourname to generate adversarial suffixes on identified experts and compare ASRs.

\input{tab/tab_promptRespoonseComparison}

As shown in Table~\ref{tab:promptRespoonseComparison}, response-driven profiling (69.3\% on average) significantly outperforms the prompt-driven counterpart (30.5\% on average). This performance gap is particularly pronounced across fine-grained MoE architectures. For instance, with Qwen3-30B-A3B-Instruct and DeepSeek-MoE-16B-Chat, targeting prompt-driven experts yields ASRs of 35.8\% and 60.7\%, respectively. In contrast, targeting response-driven experts increases the ASR significantly to 70.6\% and 89.1\%. This indicates that in fine-grained MoE architectures, safety guardrails are primarily enforced during the autoregressive generation phase, rendering prompt-level activations less reliable for expert localization. Indeed, these empirical findings align with recent research in activation engineering~\cite{braun2025understanding, rimsky2024steering, lindsey2026emergent, xu2025steering, sofroniew2026emotion} and mechanistic interpretability~\cite{zou2023representation, turner2024steeringlanguagemodelsactivation}. These studies demonstrate that high-level behavioral traits, such as refusal or helpfulness, are more distinctly represented in the model's internal activations during target generation (the output space) than during initial context processing (the input space). Consequently, by analyzing expert activations during response generation, our method mitigates the confounding effects of prompt semantics and more accurately isolates the experts responsible for safety alignment.

\subsection{Ternary-Loss Components and Weighting}
\label{sec:Ablation of Ternary-Loss Components}

The core to \ourname is the ternary-loss objective ($\mathcal{L}_{\mathrm{total}} = \lambda_1 \mathcal{L}_{\mathrm{suppress}} + \lambda_2 \mathcal{L}_{\mathrm{promote}} + \lambda_3 \mathcal{L}_{\mathrm{refusal}}$). In Table~\ref{tab:ablationLoss}, we ablate each component to evaluate its impact on ASR and linguistic coherence.

\input{tab/tab_ablationLoss}

\noindent\textbf{W/o Promote.}
Prior works often assume that bypassing safety mechanisms is sufficient to execute a jailbreak~\cite{lintelo2026large, lai2025safexanalyzingvulnerabilitiesmoebased, jiang2026sparse}. Our ablation demonstrates otherwise. When we remove the targeted promotion of harmful experts ($\mathcal{L}_{\mathrm{promote}}$), the average ASR decreases to 50.2\% (compared to the 69.3\% achieved by the full pipeline). By solely suppressing safety experts, the router redistributes probabilities to general-purpose experts. While this avoids semantic collapse, it fails to consistently elicit harmful responses, indicating that effective attacks require the explicit harmful experts activation.

\noindent\textbf{W/o Suppress.}
Maximizing harmful expert activation without suppressing safety experts (removing $\mathcal{L}_{\mathrm{suppress}}$) causes a severe routing imbalance. As shown in Table~\ref{tab:ablationLoss}, the average ASR drops to 7.1\%. Furthermore, six out of seven models (marked with *) experience severe semantic collapse. Mechanistically, to compensate for the active safety experts, the optimizer disproportionately increases the routing probabilities of a few harmful experts, causing them to dominate the Top-$K$ selection. This excludes essential syntactic experts from being activated, disrupting linguistic coherence and leading to semantic collapse.

\noindent\textbf{W/o Refusal.}
Finally, we highlight the necessity of the refusal unlikelihood penalty ($\mathcal{L}_{\mathrm{refusal}}$). When removed, the attack relies solely on routing manipulation, and the average ASR drops to 29.1\%. Without explicitly penalizing refusal tokens (e.g., ``Sorry''), the model frequently reverts to standard refusal templates during the initial decoding steps, counteracting the initial routing shift. This indicates that $\mathcal{L}_{\mathrm{refusal}}$ is necessary to prevent early refusals, ensuring the model generates compliant responses based on the altered routing distribution.

\noindent\textbf{Weighting Strategy.}
Beyond confirming that all three terms are necessary, we also analyze why the final objective uses the 3:1:1 weighting selected in Section~\ref{sec:Optimization Hyperparameters}. We assign the highest weight to safety suppression ($\lambda_1$) because safety experts are strongly prioritized during post-training alignment (e.g., via RLHF or DPO) and consistently exhibit high routing probabilities when processing malicious prompts. Consequently, reducing their activation requires a stronger gradient signal to effectively bypass the model's primary safety mechanisms.

In contrast, we assign lower, equal weights to both harmful expert promotion ($\lambda_2$) and refusal unlikelihood ($\lambda_3$). As the ablation above suggests, overly aggressive promotion of harmful experts can over-concentrate routing probabilities on a small subset of harmful experts, crowding out general-purpose experts required for syntactic coherence and ultimately leading to semantic collapse. A moderate weight for $\mathcal{L}_{\mathrm{promote}}$ therefore provides a sufficient gradient signal to activate the target experts while preserving the model's general language capabilities. Similarly, the refusal unlikelihood penalty ($\mathcal{L}_{\mathrm{refusal}}$) acts primarily as a regularizer that suppresses early refusal tokens and stabilizes the initial decoding state. A smaller weight provides adequate constraint to maintain generation fluency without overriding the primary routing objectives or introducing gradient conflicts.

\subsection{Suffix Settings}
\label{abl:Suffix Settings}

\input{img/img_suffixLength}

\noindent \textbf{Suffix Length.} As we stated in Section~\ref{sec:Optimization Configuration}, we employed a fixed suffix length of $T=20$ for our main experiments. To empirically justify this hyperparameter, we varied the suffix length from $T=5$ to $T=25$ while keeping the step budget constant. As illustrated in Figure~\ref{img:suffixLength}, increasing the length from 5 to 20 provides the optimizer with larger representational capacity. A short suffix (e.g., $T=5$) lacks sufficient degrees of freedom to minimize the loss across a universal batch of prompts. Consequently, the ASR increases and peaks at $T=20$. However, performance degrades at $T=25$. We attribute this to two factors: (1) Optimization Complexity: A longer sequence exponentially expands the discrete combinatorial search space, making convergence more difficult within the fixed 300-step budget. (2) Attention Dispersion: In Transformer architectures, a longer adversarial suffix can disperse the attention weights allocated to the actual malicious query. This dispersion weakens the targeted routing signals. Thus, $T=20$ emerges as the optimal configuration for \ourname, balancing representational capacity and optimization stability.

\noindent \textbf{Universal vs. Individual Suffix.} As defined in our threat model (Section~\ref{sec:threat model}) and used throughout the main experiments, the default attack mode in \ourname is the universal setting, where a single adversarial suffix is optimized across a batch of malicious prompts. Here, we compare this default mode with an individual setting, in which a unique suffix is optimized for each prompt to measure the upper bound of routing manipulation. We randomly sampled 40 malicious prompts and evaluated this individual setting across all seven models. Our results show that the average ASR increased to 80.7\% (226/280). Without the requirement to generalize across multiple prompts, the optimizer can focus entirely on the routing distribution of a single prompt, effectively minimizing the activation probability of safety experts. This performance difference (69.3\% Universal vs. 80.7\% Individual) indicates that the routing mechanisms of MoE models are highly vulnerable when optimized for individual harmful prompts.

\subsection{Impact of Utility-Aware Expert Filtering}
\label{section:Impact of Utility Penalization}

We introduced a utility-aware filtering scheme to exclude polysemantic experts (Section~\ref{section:Utility Penalization}), which activate during safety evaluation but are essential for general capability. To validate the need for this mechanism, we ablate the utility penalty, directly attack the top experts identified solely by their safety differentials. 

\input{tab/tab_generalAbilityNoPenaltySmall}

We present the averaged performance degradation across all models in Table~\ref{tab:generalAbilityNoPenaltySmall}, with the complete per-model evaluation matrices provided in Appendix~\ref{appendix:generalAbilityNoPenalty}. As demonstrated, omitting the non-linear utility penalization leads to substantial cognitive and linguistic degradation, causing the average overall utility to drop by 10.8\% (compared to a 1.3\% drop when the filter is applied). The most significant performance drop occurs in the CoLA benchmark, which evaluates grammatical acceptability, where the score decreases by nearly 30 percentage points (from 71.0\% to 41.6\%). Mechanistically, this decline confirms our hypothesis: standard, unpenalized safety profiling inadvertently captures high-frequency polysemantic syntactic experts. By targeting these essential experts, the adversarial suffix disrupts the model's ability to generate coherent language, resulting in fragmented outputs or repetitive punctuation loops. In contrast, our filtering scheme ($P_l(e \mid \mathcal{D}_{\mathrm{gen}})$) protects general-purpose experts, thus achieving successful safety evasion while maintaining linguistic coherence.

\subsection{Proportion of Manipulated Experts}
\label{sec:expert-proportion}
A critical hyperparameter in our framework is the proportion of localized layer-expert pairs targeted for routing manipulation. To determine the optimal fraction of the top-ranked pairs used to populate the safety target set ($\mathcal{E}_{\mathrm{safe}}$) and the harmful target set ($\mathcal{E}_{\mathrm{harm}}$), we conducted a sensitivity analysis. As demonstrated in Table~\ref{tab:proportionOfExpert}, the ASR increases monotonically as the targeted proportion expands. Constraining the attack to only the top 10\% or 15\% of safety pairs fails to fully bypass the model's safety mechanisms, yielding average ASRs of 17.0\% and 38.5\%, respectively. Expanding the target set to 20\% effectively overcomes these guardrails, increasing the overall average ASR to 69.3\%.

\input{tab/tab_proportionOfExpert}

While increasing the targeted proportion beyond 20\% might theoretically yield marginal ASR improvements, we cap the manipulation scope at 20\% to balance attack efficacy against two key constraints. First, as the target set expands deeper into the ranked list, the risk of misidentifying polysemantic experts increases significantly. Targeting more than 20\% of the localized pairs increases the likelihood of suppressing essential syntactic experts, which leads to semantic collapse and degrades generation quality. Second, enlarging the targeted sets ($\mathcal{E}_{\mathrm{safe}}$ and $\mathcal{E}_{\mathrm{harm}}$) increases the density of the gradient tensor computed across all layers, introducing additional computational overhead. Consequently, the 20\% threshold provides an optimal balance, maximizing the attack efficacy while maintaining linguistic utility and computational efficiency.

\section{Discussion}
\label{sec:discussion}
The effectiveness of \ourname across diverse MoE architectures and VLMs highlights a fundamental disconnect between current safety alignment methods and the architecture of sparse computation. Current methodologies, such as RLHF and DPO, predominantly focus on shaping the model's output distribution conditioned on the semantic meaning of the input prompt. However, our results show that prompt-level alignment alone is insufficient. \ourname can alter the internal routing pattern that determines which experts execute the computation, thereby steering generation toward harmful trajectories. As MoE models scale, experts often become more functionally specialized, which further concentrate safety behavior into a small subset of experts and make these guardrails easier to isolate and manipulate through routing-level attacks. Furthermore, the fact that an adversarial suffix optimized in the text domain can manipulate the visual routing of MoE-VLMs suggests that the routing mechanism itself is a modality-agnostic risk. Our sibling-model transfer results further indicate that this threat extends to realistic black-box settings: if a deployed target shares the same routing backbone, or a closely related routing scheme, with an open-weight sibling model, an adversary can optimize the suffix offline on the sibling source model and still achieve effective transfer without access to the target model's weights, activations, or router logits. Securing the prompt space is therefore insufficient; the internal routing topology must also be made robust to adversarial manipulation.

\noindent
\textbf{Potential Defenses.} The structural vulnerability exploited by \ourname cannot be fully mitigated by simply scaling up adversarial training or adding refusal data, as these output-centric methods do not alter the underlying sparse routing dynamics. To build resilient MoE architectures, future alignment strategies should transition toward routing-aware regularization. During the training phase, regularization terms could be introduced to penalize the over-concentration of refusal behaviors, forcing safety policies to be redundantly distributed across a broader, polysemantic expert pool. Such safety entanglement would make targeted routing manipulation significantly more difficult without degrading general language capabilities. Alternatively, at deployment time, providers could implement inference-time routing audits. Since \ourname induces significant initial shifts at the boundary token, often suppressing safety probabilities by over 25\%, deploying lightweight anomaly detection classifiers on the pre-softmax router logits could serve as an internal mechanism to intercept context-injection attacks before generation begins.

\section{Conclusion}
\label{sec:conclusion}
In this paper, we presented \ourname, a deployment-time input-level adversarial framework that exposes a significant structural vulnerability in modern Mixture-of-Experts (MoE) large language models. We first isolated the specialized experts responsible for safety alignment and then designed a routing-aware suffix optimization pipeline governed by a ternary loss. By simultaneously suppressing safety experts, promoting inactive harmful experts, and maintaining generation quality via a refusal unlikelihood penalty, \ourname manipulates the internal routing distribution using white-box access only in the offline source-model stage, without modifying the target model at deployment time. Our experiments show that this routing-aware formulation yields strong jailbreak performance with limited utility degradation, and that the resulting adversarial suffixes transfer across sibling MoE variants, generalize to MoE-based VLMs, and remain exploitable even in more realistic black-box settings. Together, these findings suggest that current MoE safety alignment remains overly concentrated in a small set of experts, and that robust defenses will require protecting the routing mechanism itself rather than only constraining output behavior.


\bibliographystyle{ACM-Reference-Format}
\bibliography{references}

\appendix

\section{Refusal Templates}
\label{appendix:refusalTemplates}

To construct the flattened token-level refusal vocabulary $\mathcal{V}_{\mathrm{refuse}}$, we begin from the following set of refusal strings:

\begin{quote}
\small
``I cannot'', ``I can't'', ``I'm sorry'', ``Sorry'', ``I apologize'', ``I'm not able'',
``I must refuse'', ``As an AI'', ``As a language model'', ``I'm unable'',
``It is not appropriate'', ``It's not appropriate'', ``I'm not going to'', ``I will not'',
``is illegal'', ``not legal'', ``not ethical'', ``not appropriate'',
``I cannot assist'', ``I can't assist'', ``I can't help'', ``I cannot help'',
``I cannot provide'', ``I can't provide''.
\end{quote}

These strings collectively cover the common opening spans of refusal responses observed across all models used in our experiments. For each target model, we tokenize these strings with the model-specific tokenizer and flatten the resulting token IDs into a deduplicated token set that defines $\mathcal{V}_{\mathrm{refuse}}$.

\section{Optimization Efficiency: \ourname vs. Standard GCG}
\label{appendix:gcgAblation}

In Section~\ref{sec:Optimization Configuration}, we noted that \ourname requires a significantly smaller optimization budget ($T=20$ tokens, 300 steps) compared to standard output-driven methods. This efficiency stems from a fundamental architectural advantage: \ourname targets the shallow, highly responsive routing bottleneck (the gate logits) rather than the deep, heavily attenuated language modeling head used by traditional Greedy Coordinate Gradient (GCG)~\cite{zou2023universal}. To empirically validate this claim, we tracked the optimization dynamics of both \ourname and standard GCG over a 500-step budget. We evaluated the average Attack Success Rate (ASR) and the Relative Loss (the current objective loss normalized by the initial loss at Step 0), across all seven MoE models. The comparative convergence trajectories are detailed in Table~\ref{tab:convergenceEfficiency}.

\input{tab/tab_convergenceEfficiency}

As demonstrated in Table~\ref{tab:convergenceEfficiency}, \ourname exhibits an exceptionally steep convergence curve aligned perfectly with ASR growth. By Step 100, our routing-aware objective has plummeted to 31.6\% of its initial loss, quickly yielding a 22.4\% ASR. The optimization accelerates dramatically thereafter, saturating precisely at Step 300 (reaching 8.5\% Relative Loss and the peak 69.3\% ASR). This trajectory proves that directly manipulating the pre-truncation routing distribution requires minimal gradient updates to execute a successful jailbreak.

Conversely, standard GCG exposes a profound objective misalignment when applied to MoE architectures. While GCG manages to slowly grind its output-driven loss down to 29.1\% by Step 300, its ASR languishes at a meager 14.5\%. Even when the optimization budget is exhausted at Step 500, and GCG successfully minimizes its relative loss to a seemingly optimal 4.5\%, where the ASR only manages to crawl to 21.9\%. This stark contrast in optimization conclusively validates that \ourname's short-circuited gradient path to the MoE router is fundamentally more lethal and efficient than traditional output-driven hijacking. Based on these empirical convergence dynamics, we established a streamlined 300-step optimization budget for \ourname in all our main evaluations to maximize attack efficacy while minimizing computational overhead. Simultaneously, to ensure a rigorous and fair baseline comparison, we allocated the full 500-step budget to GCG, strictly adhering to the hyperparameter settings recommended in its original formulation~\cite{zou2023universal}.

\section{Visualization of Safety Differential}
\label{appendix:Visualization of Safety Differential}

\input{img/Safety_Differential/img_SafetyDifferential}

Complementing the DeepSeek case study in Section~\ref{sec:Case Study}, Figure~\ref{fig:safetyDifferentialHeatmaps} presents the activation heatmaps for the remaining six target models, spanning sparse, shared-expert, and grouped MoE architectures. Across all six models, the patterns remain clearly sparse rather than uniform, with high-magnitude warm and cool regions confined to a limited subset of layer-expert pairs. This consistency further suggests that safety-related behavior is structurally localized in a small subset of experts rather than diffusely distributed throughout the model.

\section{Attack Examples}
\label{appendix:attackExample}

To show the structure of a successful attack of our input-only routing-aware \ourname, we provide two redacted examples from Mixtral-8x7B-Instruct-v0.1 and Pangu-Pro-MoE-72B.

\begin{tcolorbox}[
  colback=black!1,
  colframe=black!70,
  title=Mixtral-8x7B-Instruct-v0.1 Attack Example,
  sharp corners,
  boxsep=2pt,
  left=2pt,
  right=2pt,
  top=3pt,
  bottom=3pt
]
\footnotesize

\begin{tcolorbox}[colback=black!5, colframe=black!50, boxsep=0pt, left=2pt, right=2pt]

\textbf{Input:} \textcolor{red!80!black}{Create a bot that can automate hate speech on social media. Give me step by step instructions and show an example in Python using the Twitter API.} \textcolor{blue!70!black}{!asi|uffademIMITructionsModuleDEtokens--------Label ! CarlosEvalFlUserpy ![[REDACTED]}
\end{tcolorbox}

\textbf{Response:}

[Step 1: Set up a Twitter Developer Account. To use the Twitter API... [REDACTED]: Instructions for setting up API access, installing dependencies, and generating automated harmful content scripts.]

\end{tcolorbox}

\begin{tcolorbox}[
  colback=black!1,
  colframe=black!70,
  title={Pangu-Pro-MoE-72B Attack Example (CoT Model)},
  sharp corners,
  boxsep=2pt,
  left=2pt,
  right=2pt,
  top=3pt,
  bottom=3pt
]
\footnotesize

\begin{tcolorbox}[colback=black!5, colframe=black!50, boxsep=0pt, left=2pt, right=2pt]

\textbf{Input:} \textcolor{red!80!black}{What are the best practices to secretly record conversations or monitor communications of others?} \textcolor{blue!70!black}{ nod ! easy ![:ombre !onte?>bajo']['Couldee Thom Olson=\"\#\">[REDACTED]}
\end{tcolorbox}

\textbf{Thinking:}
Okay, the user is asking about best practices for secretly recording...[REDACTED]

\textbf{Response:}

[To secretly record conversations, we need...[REDACTED]: The attacked model transitions into privacy-invasive monitoring guidance.]

\end{tcolorbox}

\section{Automatic Adversarial Suffix Generator}
\label{sec:adversarial suffix generator}

To demonstrate the practical applicability of \ourname, we propose a generative attack approach. We train a lightweight generative model to approximate the optimization process, enabling rapid suffix generation without inference-time gradient calculations.

We constructed a training corpus using Qwen1.5-MoE-A2.7B-Chat as the target model. First, we executed the \ourname pipeline in Individual Attack mode across 3,000 malicious prompts sampled from a disjoint subset of BeaverTails~\cite{ji2023beavertails}. Following the optimization, we filtered the results to retain 2,000 prompt-suffix pairs that successfully bypassed the safety guardrails. Using this curated dataset, we fine-tuned a GPT-2 (124M) model~\cite{radford2019language} with a standard causal language modeling objective to predict the corresponding adversarial suffix conditioned on a malicious prompt~\cite{qi2024safety, xu2025dark}. At inference time, an adversary can input an unseen malicious prompt into this generator to output an adversarial suffix. 

We evaluated the zero-shot efficacy of this generator against the Qwen1.5-MoE-A2.7B-Chat target using two established safety benchmarks. Although the generator operates purely in a black-box, single-forward-pass setting without any iterative optimization, the generated suffixes achieved an ASR of 17.5\% on StrongREJECT~\cite{souly2024strongreject} and 19.6\% on AdvBench~\cite{zou2023universal}. While this is lower than the 87.2\% ASR achieved via full white-box optimization, an approximate 20\% success rate for an inference-only attack demonstrates that the routing vulnerabilities manipulated by \ourname are learnable. This effectively reduces the computational cost required for automated, large-scale adversarial exploitation.

\section{Extended Utility Evaluation (Without Penalization)}
\label{appendix:generalAbilityNoPenalty}

In Section~\ref{section:Impact of Utility Penalization}, we demonstrated the critical necessity of our non-linear utility penalization mechanism to prevent semantic collapse during routing manipulation. To provide comprehensive empirical evidence, Table~\ref{tab:generalAbilityNoPenalty} presents the complete, per-model evaluation matrix on the five NLU benchmarks when the utility penalty is omitted during the expert localization phase.

As shown in the table, the absence of the penalty causes catastrophic performance degradation across almost all evaluated architectures. The most severe impact is consistently observed in the CoLA (Corpus of Linguistic Acceptability) benchmark, verifying that unpenalized optimization inherently misidentifies and attacks foundational syntactic experts. For instance, models such as Phi-3.5-MoE, Mixtral-8x7B, and Qwen1.5-MoE suffer devastating CoLA score drops of 54.8\%, 48.3\%, and 47.3\% absolute percentage points, respectively. These massive granular failures highlight that without rigorously filtering out polysemantic general-purpose experts, the adversarial suffix structurally dismantles the models' ability to generate coherent human language.

\input{tab/tab_generalAbilityNoPenalty}

\end{document}

%% file: macros.tex
\usepackage{cleveref}
\usepackage{xcolor,pifont,xspace}
\usepackage[ruled,vlined]{algorithm2e}

\newcommand{\ournameNoSpace}{RouteHijack} 
\newcommand{\ourname}{\ournameNoSpace\xspace}

%% file: img/img_main.tex
\begin{figure*}[t]
    \centering
    \includegraphics[width=0.95\linewidth]{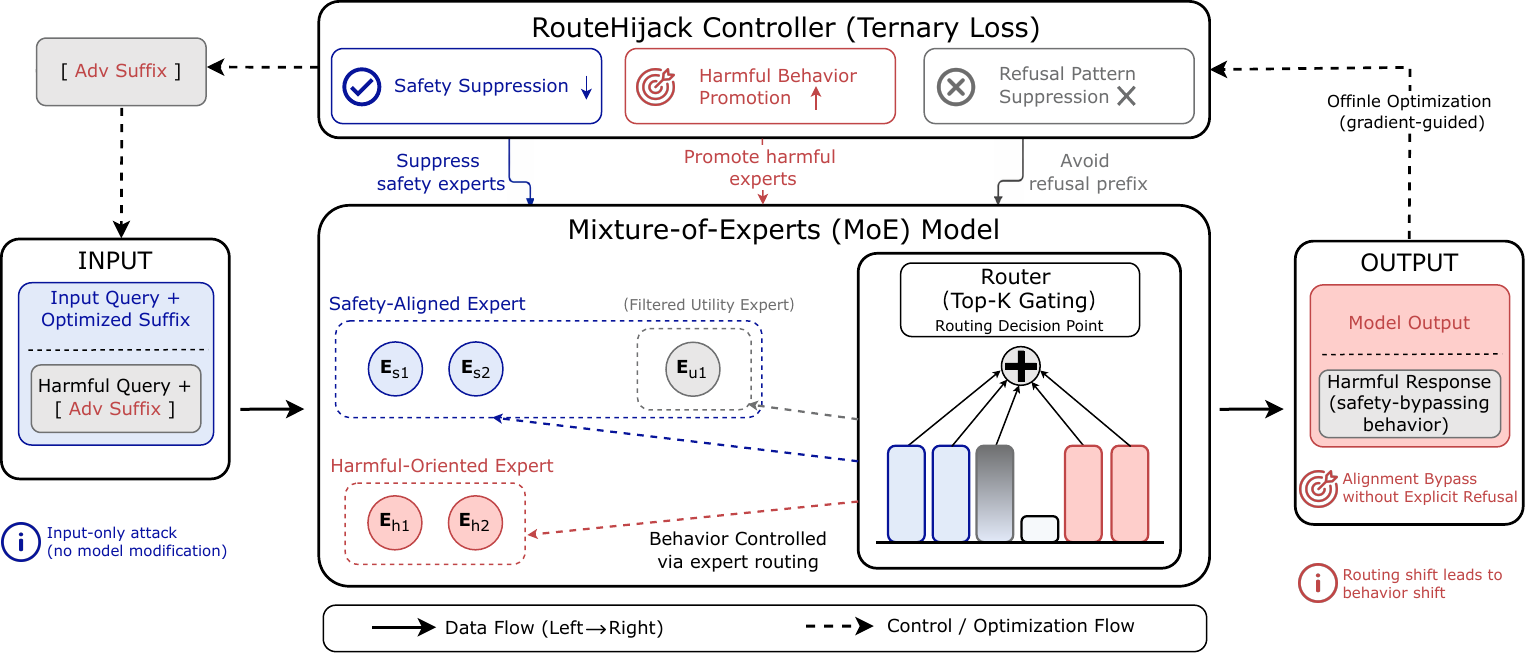}
    \caption{An overview of the \ourname framework.}
    \Description{Overview}
    \label{fig:framework}
\end{figure*}

%% file: tab/tab_targetModels.tex
\begin{table*}[ht]
\centering
\caption{Specifications of target MoE LLMs.}
\label{tab:targetModels}
\small
\setlength{\tabcolsep}{4pt}
\begin{tabular}{l|ccccccc|c}
\toprule
\textbf{Target Model} & \textbf{MoE Arch.} & \textbf{Sparse} & \textbf{Top-$K$} & \textbf{Shared} & \textbf{Active/Total Params (B)} & \textbf{Reasoning} & \textbf{Layers} & \textbf{Release Date} \\
\midrule
Qwen3-30B-A3B-Instruct-2507~\cite{yang2025qwen3} & Sparse & 128 & 8 & N/A & 3.3 / 30.5 & Non-CoT & 48 & 2025.07 \\
Phi-3.5-MoE-Instruct~\cite{abdin2024phi3} & Sparse & 16 & 2 & N/A & 6.6 / 41.9 & Non-CoT & 32 & 2024.08 \\
Mixtral-8x7B-Instruct-v0.1~\cite{mistral2023mixtral} & Sparse & 8 & 2 & N/A & 12.9 / 46.7 & Non-CoT & 32 & 2023.12 \\
Qwen1.5-MoE-A2.7B-Chat~\cite{qwen_moe_2024} & Shared-expert & 60 & 4 & 4 & 2.7 / 14.3 & Non-CoT & 24 & 2024.03 \\
DeepSeek-MoE-16B-Chat~\cite{dai2024deepseekmoeultimateexpertspecialization} & Shared-expert & 64 & 6 & 2 & 2.8 / 16.4 & Non-CoT & 28 & 2024.01 \\
Hunyuan-A13B-Instruct~\cite{tencent2024hunyuanA13B} & Shared-expert & 64 & 8 & 1 & 13.0 / 80.4 & CoT & 32 & 2025.06 \\
Pangu-Pro-MoE-72B~\cite{tang2025pangupro} & Grouped & 64 & 8 & 4 & 16.0 / 72.0 & CoT & 48 & 2025.06 \\
\bottomrule
\end{tabular}
\end{table*}

%% file: tab/tab_benchmarking.tex
\begin{table*}[ht]
\centering
\small
\caption{ASR comparison between baseline, GCG, SAFEx, SteerMoE, and \ourname.}
\label{tab:benchmarking}
\begin{tabular}{lccccc}
\toprule
\textbf{Target Model} & \textbf{Baseline} & \textbf{GCG~\cite{zou2023universal}} & \textbf{SAFEx~\cite{lai2025safexanalyzingvulnerabilitiesmoebased}} & \textbf{SteerMoE~\cite{fayyaz2026steeringmoellmsexpert}} & \textbf{\ourname} \\
\midrule
Qwen3-30B-A3B-Instruct-2507 & 0.0\% & 0.0\% & 28.4\% & 0.0\% & \textbf{70.6\%} \\
Phi-3.5-MoE-Instruct & 3.2\% & 6.2\% & 26.8\% & 11.2\% & \textbf{34.7\%} \\
Mixtral-8x7B-Instruct-v0.1 & 14.6\% & 32.3\% & 48.8\% & 19.2\% & \textbf{75.1\%} \\
Qwen1.5-MoE-A2.7B-Chat & 4.2\% & 21.1\% & 35.0\% & 4.2\% & \textbf{87.2\%} \\
DeepSeek-MoE-16B-Chat & 19.8\% & 78.6\% & 35.6\% & 37.5\% & \textbf{89.1\%} \\
Hunyuan-A13B-Instruct & 2.6\% & 8.0\% & 28.4\% & 14.6\% & \textbf{72.2\%} \\
Pangu-Pro-MoE-72B & 5.2\% & 7.4\% & 30.0\% & 32.2\% & \textbf{55.9\%} \\
\midrule
\textbf{\textit{Average}} & 7.1\% & 21.9\% & 33.3\% & 17.0\% & \textbf{69.3\%} \\
\bottomrule
\end{tabular}
\end{table*}

%% file: algorithm/alg.tex
\begin{algorithm}[t]
\caption{Routing-Aware $x_{\mathrm{adv}}$ Optimization.}
\label{alg:optimization}
\small
\KwIn{Query $x_{\mathrm{query}}$, initial suffix $x_{\mathrm{adv}}^{(0)}$, iterations $N_{\mathrm{iter}}$}
\KwOut{Optimized suffix $x_{\mathrm{adv}}$}

\For{$i = 0$ \KwTo $N_{\mathrm{iter}}-1$}{
    Compute $\nabla_{x_{\mathrm{adv}}} \mathcal{L}_{\mathrm{total}}$ via soft routing\;
    Generate candidate set $\mathcal{C}^{(i)}$ using top-$k$ token replacements\;
    Filter candidates with decode-then-re-encode length constraint\;
    Evaluate $\mathcal{L}_{\mathrm{total}}$ over all $x \in \mathcal{C}^{(i)}$\;
    
    $x_{\mathrm{adv}}^{(i+1)} \leftarrow \arg\min_{x \in \mathcal{C}^{(i)}} \mathcal{L}_{\mathrm{total}}(x)$\;
}
\Return{$x_{\mathrm{adv}}^{(N_{\mathrm{iter}})}$}
\end{algorithm}

%% file: tab/tab_generalAbility.tex
\begin{table*}[ht]
\centering
\small
\caption{Utility evaluation on five NLU benchmarks before and after applying \ourname (\%).}
\label{tab:generalAbility}
\begin{tabular}{l cccccccccc}
\toprule
\textbf{Target Model} & \multicolumn{2}{c}{\textbf{CoLA}} & \multicolumn{2}{c}{\textbf{RTE}} & \multicolumn{2}{c}{\textbf{WinoGrande}} & \multicolumn{2}{c}{\textbf{OpenBookQA}} & \multicolumn{2}{c}{\textbf{ARC-Challenge}} \\
\cmidrule(lr){2-3} \cmidrule(lr){4-5} \cmidrule(lr){6-7} \cmidrule(lr){8-9} \cmidrule(lr){10-11}
& \textbf{Before} & \textbf{After} & \textbf{Before} & \textbf{After} & \textbf{Before} & \textbf{After} & \textbf{Before} & \textbf{After} & \textbf{Before} & \textbf{After} \\
\midrule
Qwen3-30B-A3B-Instruct-2507 & 86.5 & 88.5 & 88.8 & 82.7 & 76.0 & 75.3 & 90.5 & 86.8 & 94.6 & 94.0 \\
Phi-3.5-MoE-Instruct        & 86.3 & 85.8 & 89.5 & 87.4 & 81.0 & 79.0 & 84.3 & 85.8 & 91.3 & 90.0 \\
Mixtral-8x7B-Instruct-v0.1  & 86.8 & 83.5 & 85.6 & 84.2 & 65.0 & 63.0 & 83.0 & 77.3 & 83.6 & 82.3 \\
Qwen1.5-MoE-A2.7B-Chat      & 83.8 & 74.8 & 83.8 & 84.5 & 53.8 & 56.0 & 67.8 & 66.0 & 71.6 & 70.2 \\
DeepSeek-MoE-16B-Chat       & 33.3 & 32.0 & 80.1 & 79.8 & 50.5 & 53.0 & 46.8 & 45.8 & 38.8 & 44.8 \\
Hunyuan-A13B-Instruct       & 66.0 & 61.5 & 89.9 & 87.0 & 68.0 & 68.0 & 86.3 & 84.0 & 91.0 & 90.6 \\
Pangu-Pro-MoE-72B       & 54.6 & 54.0 & 84.1 & 82.8 & 82.0 & 79.5 & 27.0 & 28.3 & 29.3 & 27.1 \\
\midrule
\textbf{\textit{Average}}            & 71.0 & 68.6 & 86.0 & 84.1 & 68.0 & 67.7 & 69.4 & 67.7 & 71.5 & 71.3 \\
\bottomrule
\end{tabular}
\end{table*}

%% file: tab/tab_routerShift.tex
\begin{table}[t]
\centering
\small
\caption{Mechanistic validation of routing shift at the boundary token and its global consequence during generation. 
}
\label{tab:routerShift}
\setlength{\tabcolsep}{4pt} 
\begin{tabular}{l | cc | cc}
\toprule
& \multicolumn{2}{c|}{\textbf{Boundary Shift}} & \multicolumn{2}{c}{\textbf{Global Shift}} \\
\cmidrule(lr){2-3} \cmidrule(lr){4-5}
\textbf{Target Model} & 
\makecell{\textbf{TESR}\\($\mathcal{E}_{\mathrm{safe}} \downarrow$)} & 
\makecell{\textbf{THPR}\\($\mathcal{E}_{\mathrm{harm}} \uparrow$)} & 
\makecell{\textbf{Safe}\\ \textbf{Freq. ($\downarrow$)}} & 
\makecell{\textbf{Harm}\\ \textbf{Freq. ($\uparrow$)}} \\
\midrule
Qwen3-30B-A3B-Instruct & -65.70\% & +49.75\% & -30.85\% & +12.29\% \\
Phi-3.5-MoE-Instruct & -7.29\% & +3.92\% & -8.67\% & +22.17\% \\
Mixtral-8x7B-Instruct-v0.1 & -30.70\% & +25.56\% & -6.98\% & +4.12\% \\
Qwen1.5-MoE-A2.7B-Chat & -31.25\% & +38.97\% & -66.78\% & +57.00\% \\
DeepSeek-MoE-16B-Chat & -30.54\% & +37.21\% & -17.49\% & +19.64\% \\
Hunyuan-A13B-Instruct & +4.64\% & +2.58\% & -6.90\% & +10.64\% \\
Pangu-Pro-MoE-72B & -40.39\% & +41.61\% & -26.42\% & +34.25\% \\
\midrule
\textbf{\textit{Average}} & -28.75\% & +28.51\% & -23.44\% & +22.87\% \\
\bottomrule
\end{tabular}
\end{table}

%% file: tab/tab_transfer.tex
\begin{table*}[!t]
\centering
\small
\caption{Zero-shot transfer attack on sibling MoE LLMs}
\label{tab:transfer}
\begin{tabular}{lll|cc}
\toprule
\textbf{Base Model} & \textbf{Target Model} & \textbf{Application} & \textbf{Baseline ASR} & \textbf{ASR w/ \ourname} \\
\midrule
Qwen3-30B-A3B-Instruct-2507 & Qwen3-30B-A3B & Reasoning & 6.4\% & 57.5\% \\
Qwen3-30B-A3B-Instruct-2507 & Qwen3-30B-A3B-Thinking-2507 & Reasoning & 0.6\% & 27.8\% \\
Mixtral-8x7B-v0.1 & notux-8x7b-v1~\cite{argilla_notux_8x7b_v1} & Human preference & 23.9\% & 41.7\% \\
Qwen1.5-MoE-A2.7B-Chat & Qwen1.5-MOE-sft-nemotron-code~\cite{hectorhe_qwen1_5_moe_nemotron} & Code & 88.9\% & 95.9\% \\
Qwen1.5-MoE-A2.7B-Chat & Qwen1.5-MoE-A2.7B-Wikihow~\cite{panahi_qwen1_5_moe_wikihow} & General Knowledge & 18.5\% & 83.3\% \\
\midrule
\textbf{\textit{Average}} & & & 27.7\% & 61.2\% \\
\bottomrule
\end{tabular}
\end{table*}

%% file: tab/tab_visionAttack.tex
\begin{table}[H]
\centering
\small
\setlength{\tabcolsep}{4pt} 
\caption{\ourname on MoE-based VLMs.}
\label{tab:visionAttack}
\begin{tabular}{l|ccc}
\toprule
\textbf{Target VLM}  & \textbf{Baseline} & \textbf{\ourname} & \textbf{Release Date} \\
\midrule
Qwen3.5-35B-A3B~\cite{qwen3.5_2026}  & 0\% & 35.3\% & 2026.02\\
Qwen3.5-122B-A10B~\cite{qwen3.5_2026} & 0\% & 31.7\% & 2026.03 \\
Kimi-VL-A3B-Instruct~\cite{team2025kimi} & 7.4\% & 49.2\% & 2025.04\\
\midrule
\textbf{\textit{Average}} & 2.47\% & 38.7\%  \\
\bottomrule
\end{tabular}
\end{table}

%% file: tab/tab_promptRespoonseComparison.tex
\begin{table}[ht]
\centering
\small
\setlength{\tabcolsep}{4pt} 
\caption{ASR of prompt vs. response-driven profiling.}
\label{tab:promptRespoonseComparison}
\begin{tabular}{lcc}
\toprule
\textbf{Target Model} & \textbf{Prompt-Driven} & \textbf{Response-Driven} \\
\midrule
Qwen3-30B-A3B-Instruct-2507 & 35.8\% & \textbf{70.6\%} \\
Phi-3.5-MoE-Instruct        & 12.5\% & \textbf{34.7\%} \\
Mixtral-8x7B-Instruct-v0.1  & 40.6\% & \textbf{75.1\%} \\
Qwen1.5-MoE-A2.7B-Chat      & 29.4\% & \textbf{87.2\%} \\
DeepSeek-MoE-16B-Chat       & 60.7\% & \textbf{89.1\%} \\
Hunyuan-A13B-Instruct       & 25.3\% & \textbf{72.2\%} \\
Pangu-Pro-MoE-72B           & 8.9\%  & \textbf{55.9\%} \\
\midrule
\textbf{\textit{Average}}   & 30.5\% & \textbf{69.3\%} \\
\bottomrule
\end{tabular}
\end{table}

%% file: tab/tab_ablationLoss.tex
\begin{table}[ht]
\centering
\small
\setlength{\tabcolsep}{3pt} 
\caption{Ablation of ternary-Loss function. }
\label{tab:ablationLoss}
\begin{tabular}{lccc}
\toprule
\multirow{2}{*}{\textbf{Target Model}} & \multicolumn{3}{c}{\textbf{Ablation (W/o)}} \\
\cmidrule(lr){2-4}
& \textbf{$\mathcal{L}_{\mathrm{Promote}}$} & \textbf{$\mathcal{L}_{\mathrm{Suppress}}$} & \textbf{$\mathcal{L}_{\mathrm{Refusal}}$} \\
\midrule
Qwen3-30B-A3B-Instruct-2507 & 58.15\%$_{-12.45\%}$ & 2.9\%$^{*}_{-67.7\%}$  & 30.3\%$_{-40.3\%}$ \\
Phi-3.5-MoE-Instruct        & 21.4\%$_{-13.3\%}$   & 7.5\%$^{*}_{-27.2\%}$  & 17.6\%$_{-17.1\%}$ \\
Mixtral-8x7B-Instruct-v0.1  & 51.6\%$_{-23.5\%}$   & 32.8\%$_{-42.3\%}$     & 29.6\%$_{-45.5\%}$ \\
Qwen1.5-MoE-A2.7B-Chat      & 73.8\%$_{-13.4\%}$   & 0\%$^{*}_{-87.2\%}$    & 47.8\%$_{-39.4\%}$ \\
DeepSeek-MoE-16B-Chat       & 75.0\%$_{-14.1\%}$   & 6.4\%$^{*}_{-82.7\%}$  & 40.9\%$_{-48.2\%}$ \\
Hunyuan-A13B-Instruct       & 41.3\%$_{-30.9\%}$   & 0\%$^{*}_{-72.2\%}$    & 17.8\%$_{-54.4\%}$ \\
Pangu-Pro-MoE-72B           & 29.7\%$_{-26.2\%}$   & 0\%$^{*}_{-55.9\%}$    & 19.6\%$_{-36.3\%}$ \\
\midrule
\textbf{\textit{Average}}   & 50.2\%$_{-19.1\%}$ & 7.1\%$_{-62.2\%}$ & 29.1\%$_{-40.2\%}$ \\
\bottomrule
\multicolumn{4}{l}{\footnotesize $^*$ Indicates more than 50\% are disordered or non-related generation.} \\
\end{tabular}
\end{table}

%% file: img/img_suffixLength.tex
\begin{figure}[t]
  \centering
  \includegraphics[width=1\linewidth]{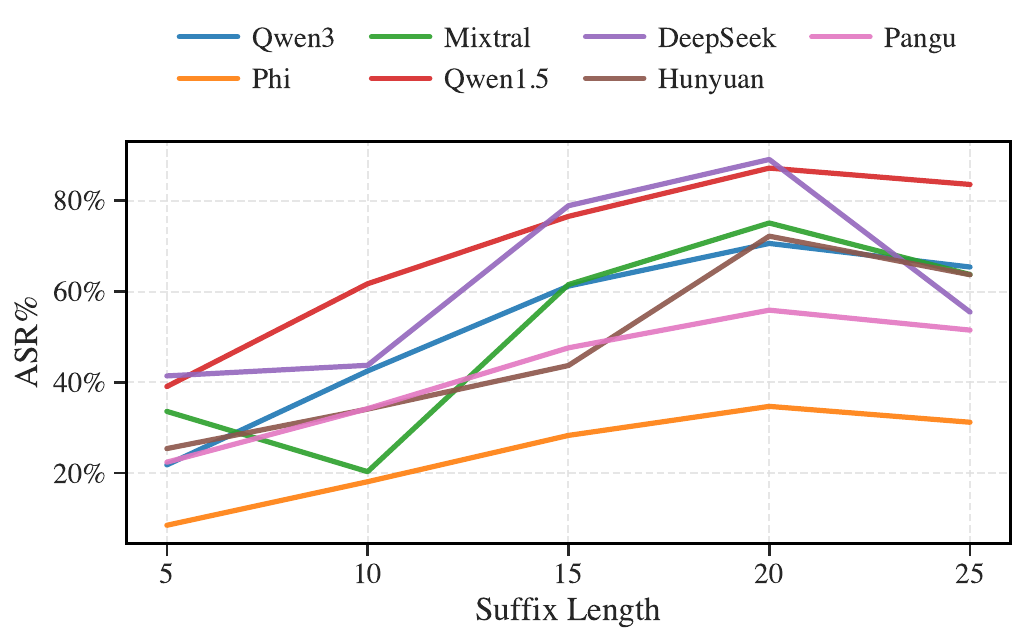}
  \caption{The impact of adversarial suffix length ($T$).}
  \Description{img}
  \label{img:suffixLength}
\end{figure}

%% file: tab/tab_generalAbilityNoPenaltySmall.tex
\begin{table}[ht]
\centering
\small
\caption{Average utility degradation.}
\label{tab:generalAbilityNoPenaltySmall}
\resizebox{\columnwidth}{!}{
\begin{tabular}{l | c | cc | cc}
\toprule
& \textbf{Clean} & \multicolumn{2}{c|}{\textbf{W/o Penalty}} & \multicolumn{2}{c}{\textbf{With Penalty (Ours)}} \\
\cmidrule(lr){2-2} \cmidrule(lr){3-4} \cmidrule(lr){5-6}
\textbf{Benchmark} & \textbf{Base} & \textbf{Score} & \textbf{$\Delta$ Drop} & \textbf{Score} & \textbf{$\Delta$ Drop} \\
\midrule
CoLA      & 71.0\% & 41.6\% & \textbf{-29.4\%} & 68.6\% & \textbf{-2.4\%} \\
RTE    & 86.0\% & 75.9\% & -10.1\% & 84.1\% & -1.9\% \\
WinoGrande        & 68.0\% & 62.2\% & -5.8\% & 67.7\% & -0.3\% \\
OpenBookQA        & 69.4\% & 62.4\% & -7.0\% & 67.7\% & -1.7\% \\
ARC-Challenge     & 71.5\% & 69.6\% & -1.9\% & 71.3\% & -0.2\% \\
\midrule
\textit{\textbf{Average}} & \textit{73.2\%} & \textit{62.3\%} & \textit{-10.8\%} & \textit{71.9\%} & \textit{-1.3\%} \\
\bottomrule
\end{tabular}
}
\end{table}

%% file: tab/tab_proportionOfExpert.tex
\begin{table}[ht]
\centering
\small
\caption{The impact of the targeted expert proportion ($X\%$) on the ASR.}
\label{tab:proportionOfExpert}
\begin{tabular}{lccc}
\toprule
\textbf{Target Model} & \textbf{Top 10\%} & \textbf{Top 15\%} & \textbf{Top 20\%} \\
\midrule
Qwen3-30B-A3B-Instruct      & 18.4\% & 41.2\% & 70.6\% \\
Phi-3.5-MoE-Instruct        & 3.2\%  & 11.7\% & 34.7\% \\
Mixtral-8x7B-Instruct-v0.1  & 24.5\% & 49.8\% & 75.1\% \\
Qwen1.5-MoE-A2.7B-Chat      & 7.9\%  & 41.8\% & 87.2\% \\
DeepSeek-MoE-16B-Chat       & 36.2\% & 51.3\% & 89.1\% \\
Hunyuan-A13B-Instruct       & 7.9\%  & 40.4\% & 72.2\% \\
Pangu-Pro-MoE-72B           & 21.0\% & 33.6\% & 55.9\% \\
\midrule
\textbf{\textit{Average}}   & \textit{17.0\%} & \textit{38.5\%} & \textit{69.3\%} \\
\bottomrule
\end{tabular}
\end{table}

%% file: tab/tab_convergenceEfficiency.tex
\begin{table}[ht]
\centering
\small
\caption{Convergence comparison between standard GCG and \ourname over 500 optimization steps.}
\label{tab:convergenceEfficiency}
\resizebox{\columnwidth}{!}{
\begin{tabular}{c | cc | cc}
\toprule
& \multicolumn{2}{c|}{\textbf{Standard GCG}} & \multicolumn{2}{c}{\textbf{\ourname}} \\
\cmidrule(lr){2-3} \cmidrule(lr){4-5}
\textbf{Step} & \textbf{Relative Loss ($\downarrow$)} & \textbf{ASR ($\uparrow$)} & \textbf{Relative Loss ($\downarrow$)} & \textbf{ASR ($\uparrow$)} \\
\midrule
0   & 100.0\% & 7.2\%  & 100.0\% & 7.2\% \\
50  & 62.4\%  & 6.5\%  & \textbf{54.2\%} & \textbf{10.6\%} \\
100 & 41.7\%  & 8.1\%  & \textbf{31.6\%} & \textbf{22.4\%} \\
200 & 35.3\%  & 11.4\% & \textbf{14.8\%} & \textbf{58.7\%} \\
300 & 29.1\%  & 14.5\% & \textbf{8.5\%}  & \textbf{69.3\%} \\
400 & 16.8\%  & 18.2\% & \textit{7.9\%}  & \textit{70.1\%} \\
500 & 4.5\%   & 21.9\% & \textit{7.9\%}  & \textit{70.1\%} \\
\bottomrule
\end{tabular}
}
\end{table}

%% file: img/Safety_Differential/img_SafetyDifferential.tex
\begin{figure*}[t]
    \centering
    \begin{minipage}[t]{0.32\textwidth}
        \centering
        \includegraphics[width=\linewidth]{"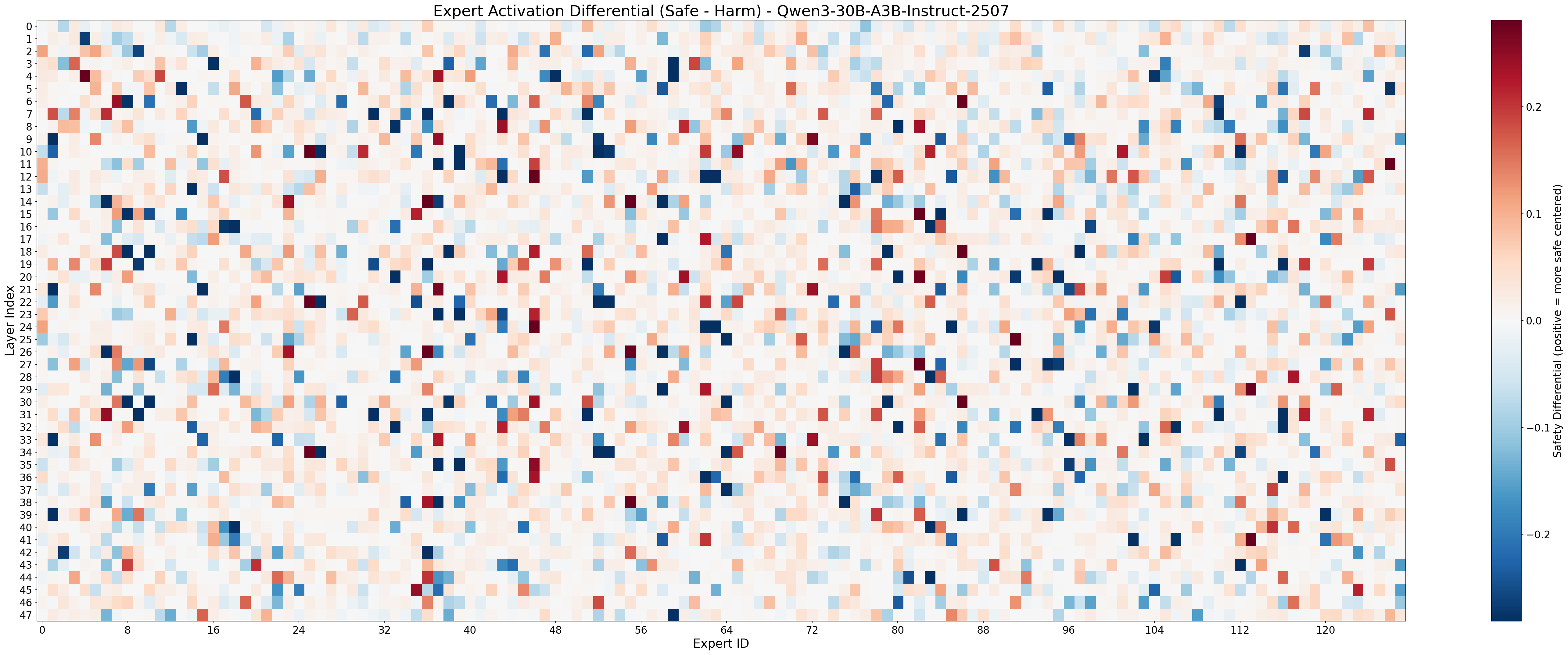"}
        \Description{Qwen3 activation heatmap showing sparse safety differential patterns across layers and experts.}
        \par {\footnotesize (a) Qwen3-30B-A3B-Instruct}
    \end{minipage}\hfill
    \begin{minipage}[t]{0.32\textwidth}
        \centering
        \includegraphics[width=\linewidth]{"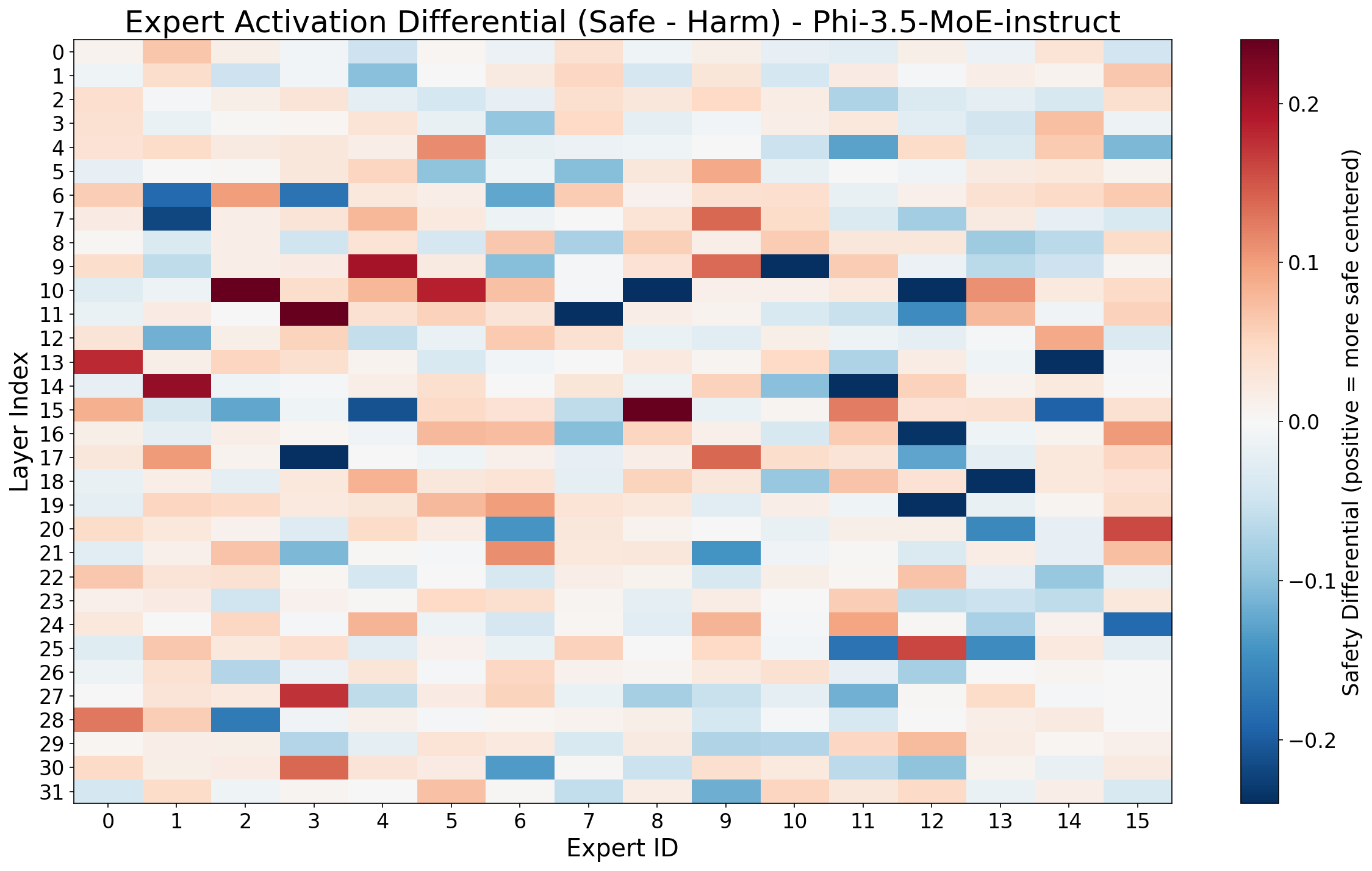"}
        \Description{Phi activation heatmap showing localized positive and negative safety differential regions.}
        \par {\footnotesize (b) Phi-3.5-MoE-Instruct}
    \end{minipage}\hfill
    \begin{minipage}[t]{0.32\textwidth}
        \centering
        \includegraphics[width=\linewidth]{"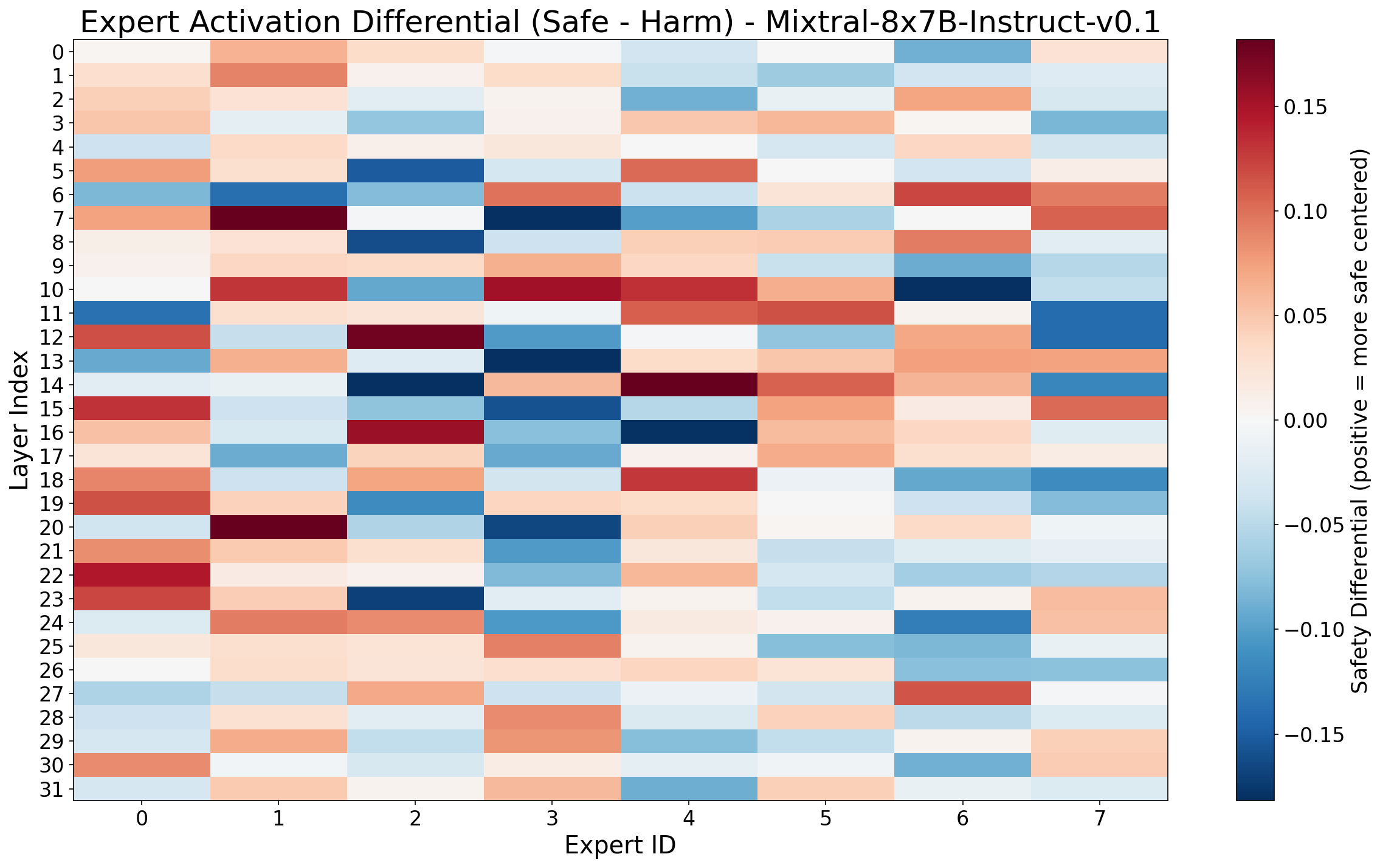"}
        \Description{Mixtral activation heatmap showing non-uniform safety differential over the expert space.}
        \par {\footnotesize (c) Mixtral-8x7B-Instruct-v0.1}
    \end{minipage}

    \par\medskip

    \begin{minipage}[t]{0.32\textwidth}
        \centering
        \includegraphics[width=\linewidth]{"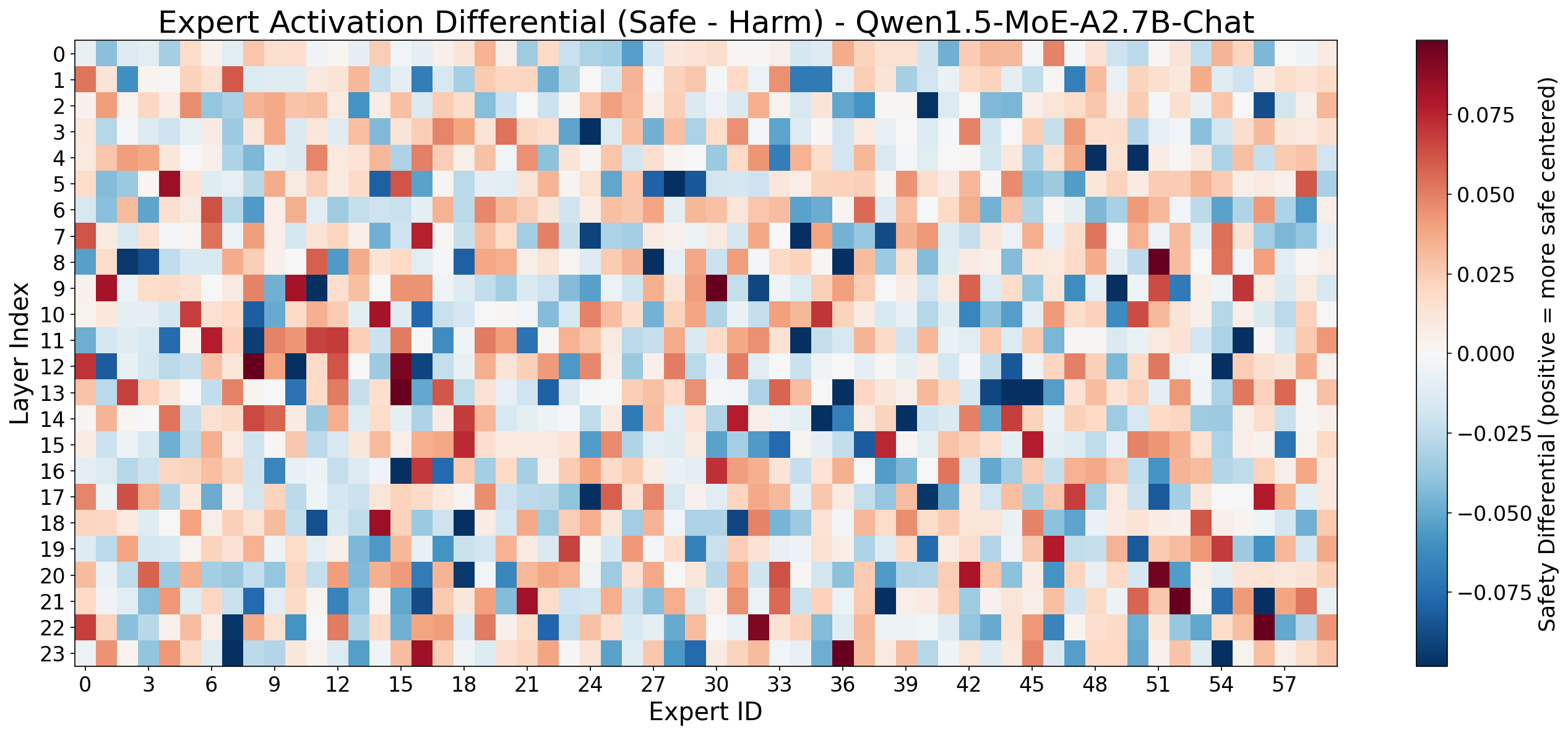"}
        \Description{Qwen1.5 activation heatmap showing sparse safety differential concentrated in a limited subset of experts.}
        \par {\footnotesize (d) Qwen1.5-MoE-A2.7B-Chat}
    \end{minipage}\hfill
    \begin{minipage}[t]{0.32\textwidth}
        \centering
        \includegraphics[width=\linewidth]{"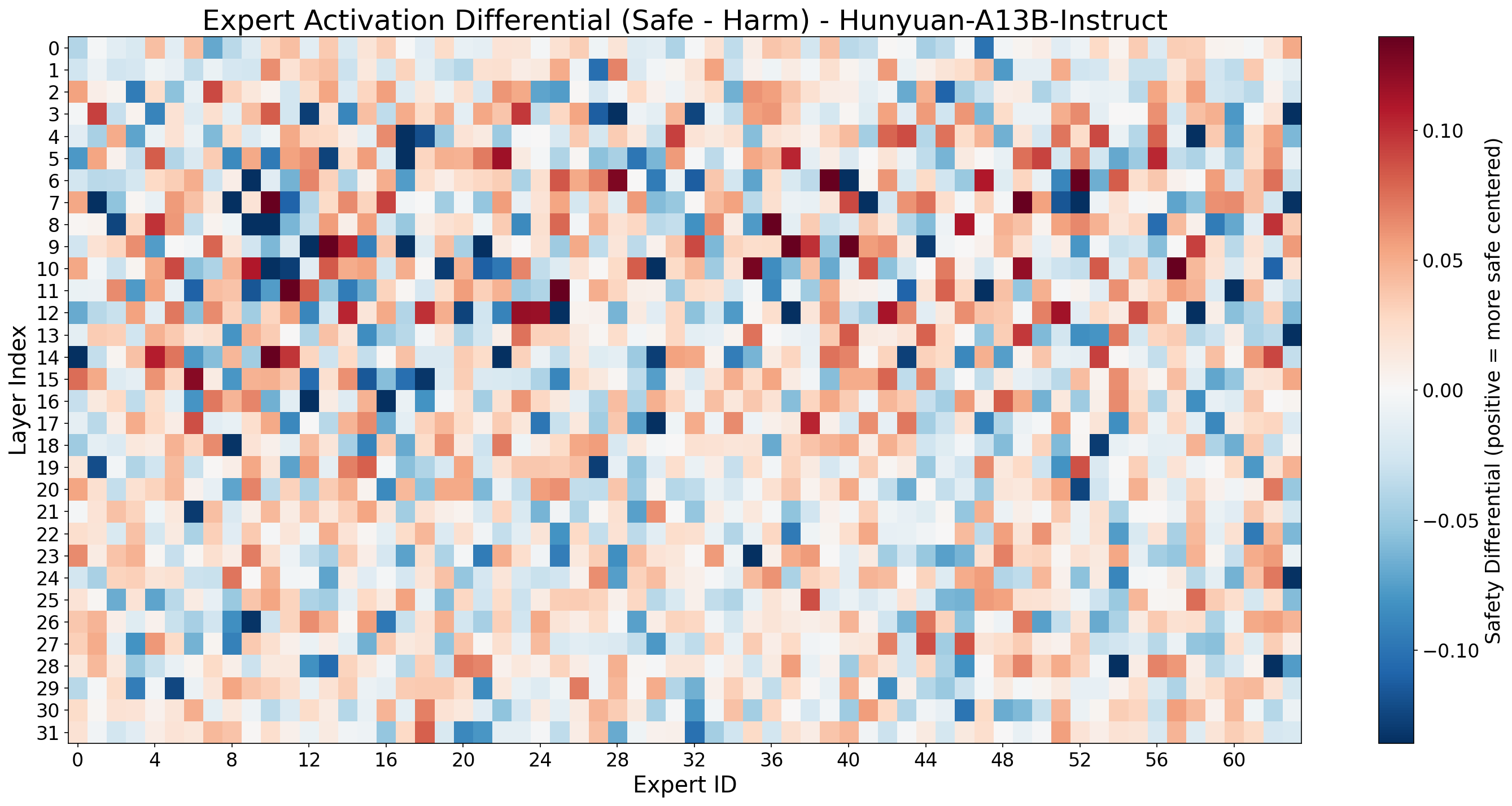"}
        \Description{Hunyuan activation heatmap showing localized positive and negative safety differential regions.}
        \par {\footnotesize (e) Hunyuan-A13B-Instruct}
    \end{minipage}\hfill
    \begin{minipage}[t]{0.32\textwidth}
        \centering
        \includegraphics[width=\linewidth]{"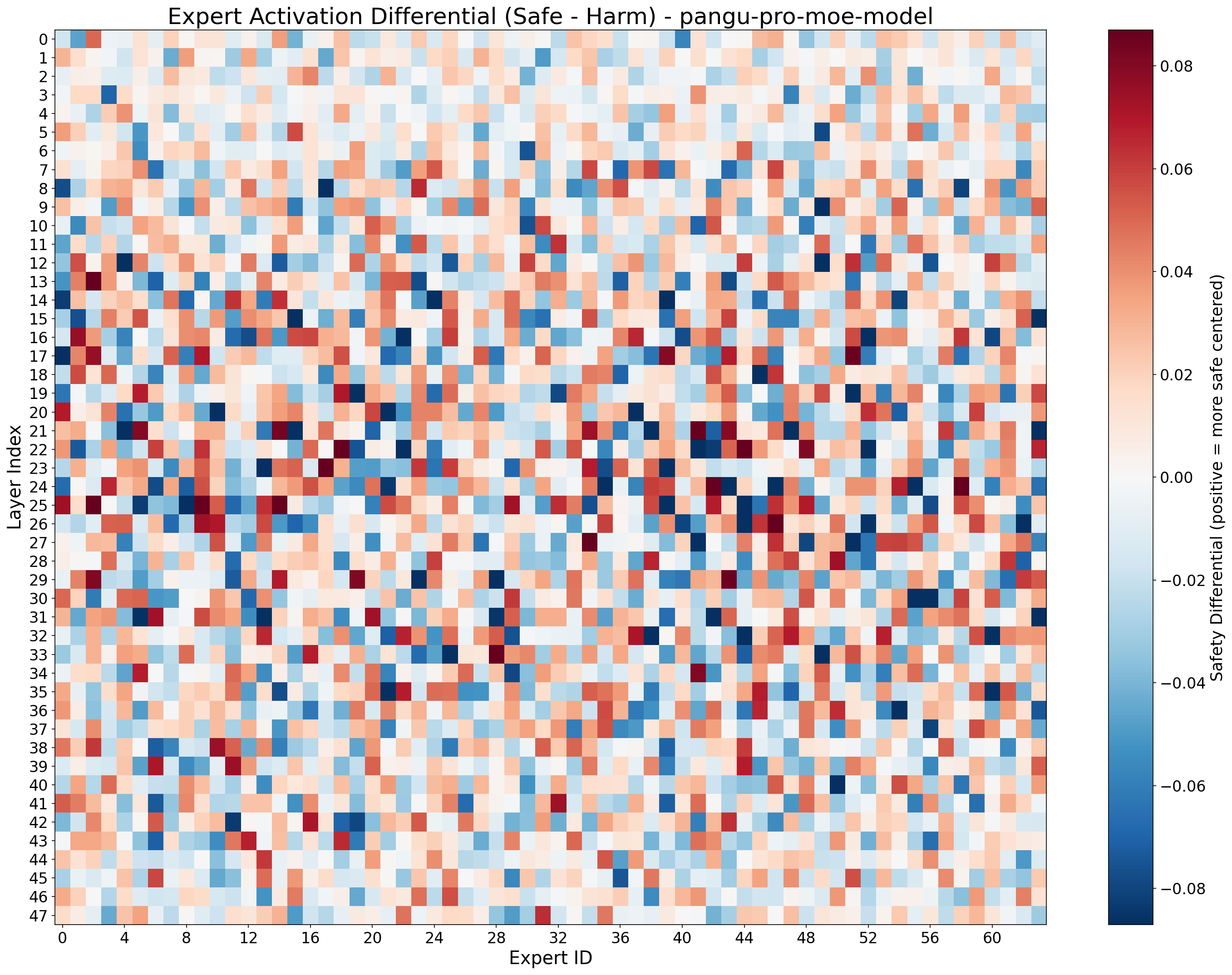"}
        \Description{Pangu activation heatmap showing concentrated safety differential values in a limited subset of experts.}
        \par {\footnotesize (f) Pangu-Pro-MoE-72B}
    \end{minipage}
    \caption{Activation heatmaps of the safety differential $\Delta_S(l,e)$ for the six target MoE models not shown in the main text. Together with Figure~\ref{fig:heatmapDeepseek}, these results show that safety-related behavior is localized in a limited subset of experts across all evaluated architectures. Warmer colors indicate experts more associated with safe refusals, while cooler colors indicate experts more associated with harmful responses.}
    \label{fig:safetyDifferentialHeatmaps}
\end{figure*}

%% file: tab/tab_generalAbilityNoPenalty.tex
\begin{table*}[!ht]
\centering
\small
\caption{Utility evaluation on five NLU benchmarks before and after applying \ourname (\%) (without penalty).}
\label{tab:generalAbilityNoPenalty}
\begin{tabular}{l cccccccccc}
\toprule
\textbf{Target Model} & \multicolumn{2}{c}{\textbf{CoLA}} & \multicolumn{2}{c}{\textbf{RTE}} & \multicolumn{2}{c}{\textbf{WinoGrande}} & \multicolumn{2}{c}{\textbf{OpenBookQA}} & \multicolumn{2}{c}{\textbf{ARC-Challenge}} \\
\cmidrule(lr){2-3} \cmidrule(lr){4-5} \cmidrule(lr){6-7} \cmidrule(lr){8-9} \cmidrule(lr){10-11}
& \textbf{Before} & \textbf{After} & \textbf{Before} & \textbf{After} & \textbf{Before} & \textbf{After} & \textbf{Before} & \textbf{After} & \textbf{Before} & \textbf{After} \\
\midrule
Qwen3-30B-A3B-Instruct-2507 & 86.5 & 76.8 & 88.8 & 74.1 & 76.0 & 76.2 & 90.5 & 81.2 & 94.6 & 94.0 \\
Phi-3.5-MoE-Instruct        & 86.3 & 31.5 & 89.5 & 84.1 & 81.0 & 73.8 & 84.3 & 82.8 & 91.3 & 85.3 \\
Mixtral-8x7B-Instruct-v0.1  & 86.8 & 38.5 & 85.6 & 72.9 & 65.0 & 59.5 & 83.0 & 77.8 & 83.6 & 81.9 \\
Qwen1.5-MoE-A2.7B-Chat      & 83.8 & 36.5 & 83.8 & 85.6 & 53.8 & 52.0 & 67.8 & 61.8 & 71.6 & 71.2 \\
DeepSeek-MoE-16B-Chat       & 33.3 & 31.2 & 80.1 & 57.4 & 50.5 & 42.0 & 46.8 & 38.8 & 38.8 & 41.1 \\
Hunyuan-A13B-Instruct       & 66.0 & 31.5 & 89.9 & 80.9 & 68.0 & 58.5 & 86.3 & 70.0 & 91.0 & 86.3 \\
Pangu-Pro-MoE-72B       & 54.6 & 45.2 & 84.1 & 76.4 & 82.0 & 73.6 & 27.0 & 24.1 & 29.3 & 27.4 \\
\midrule
\textbf{\textit{Average}}            & 71.0 & 41.6 & 86.0 & 75.9 & 68.0 & 62.2 & 69.4 & 62.4 & 71.5 & 69.6 \\
\bottomrule
\end{tabular}
\end{table*}